%% file: main.tex
\documentclass[11pt]{article}

\usepackage[final]{acl}

\usepackage{times}
\usepackage{latexsym}
\usepackage{placeins}

\usepackage[T1]{fontenc}

\usepackage[utf8]{inputenc}

\usepackage{microtype}

\usepackage{inconsolata}

\usepackage{graphicx}
\usepackage{booktabs}
\usepackage{multirow}
\usepackage{amsmath}
\usepackage{xcolor}
\usepackage{caption}
\usepackage{amssymb}
\usepackage{subfigure}
\usepackage{subcaption}
\usepackage{color}
\usepackage[table]{xcolor}
\usepackage{hyperref}
\usepackage{url}
\usepackage{enumitem}
\usepackage{arydshln}

\definecolor{lightgreen}{RGB}{220,245,220}
\definecolor{lightorange}{RGB}{255,235,215}
\newcommand{\good}[1]{\colorbox{lightgreen}{#1}}
\newcommand{\bad}[1]{\colorbox{lightorange}{#1}}

\title{MetaMem: Evolving Meta-Memory for Knowledge Utilization through Self-Reflective Symbolic Optimization}

\author{Haidong Xin$^{1}$\thanks{ \ \ indicates equal contribution.}, Xinze Li$^{1}$\footnotemark[1], Zhenghao Liu$^{1}$\thanks{ \ \ indicates corresponding author.}, Yukun Yan$^{2}\footnotemark[2]$,\\ 
\textbf{Shuo Wang$^{2}$, Cheng Yang$^{3}$, Yu Gu$^{1}$, Ge Yu$^{1}$ and Maosong Sun$^{2}$} \\ 
$^1$School of Computer Science and Engineering, Northeastern University, Shenyang, China \\
$^2$Department of Computer Science and Technology, Tsinghua University, Beijing, China \\
$^3$School of Computer Science, Beijing University of Posts and Telecommunications,\\
Beijing, China\\
}

\def\method{MetaMem}

\begin{document}
\maketitle
\input{section/0_abstract}
\input{section/1_intro}
\input{section/2_related}
\input{section/3_methodology}
\input{section/4_experiment}

\input{section/5_results}
\input{section/6_conclusion}

\section*{Acknowledgments}
This work is supported by the National Natural Science Foundation of China (No. 62576082). This work is also supported by the AI9Stars community.

\input{section/7_limitations}


\bibliography{custom}

\clearpage
\appendix
\input{section/8_appendix}

\end{document}

%% file: section/0_abstract.tex
\begin{abstract}
Existing memory systems enable Large Language Models (LLMs) to support long-horizon human-LLM interactions by persisting historical interactions beyond limited context windows.
However, while recent approaches have succeeded in constructing effective memories, they often disrupt the inherent logical and temporal relationships within interaction sessions, resulting in fragmented memory units and degraded reasoning performance.
In this paper, we propose \method{}, a novel framework that augments memory systems with a self-evolving meta-memory, aiming to teach LLMs how to effectively utilize memorized knowledge.
During meta-memory optimization, \method{} iteratively distills transferable knowledge utilization experiences across different tasks by self-reflecting on reasoning processes and performing actions to update the current meta-memory state.
The accumulated meta-memory units serve as explicit knowledge utilization experiences, guiding the LLM to systematically identify and integrate critical evidence from scattered memory fragments.
Extensive experiments demonstrate the effectiveness of \method{}, which significantly outperforms strong baselines by over 3.6\%.
All codes and datasets are available at \url{https://github.com/OpenBMB/MetaMem}.
\end{abstract}

%% file: section/1_intro.tex
\input{figure/intro}
\section{Introduction}
Large Language Models (LLMs)~\cite{liu2024deepseek,yang2025qwen3,team2025kimi} have demonstrated remarkable capabilities in complex tasks~\cite{lewis2020retrieval,he2021efficient,liu5459034knowledge}, largely attributed to their emergent abilities.
Nevertheless, LLMs are usually constrained by limited context window sizes~\cite{wang2024beyond,liu2025comprehensive}, which restricts their effectiveness in handling long-horizon historical human-LLM interaction scenarios.
To address this limitation, researchers have developed external memory systems to persist historical context beyond the model's context window~\cite{packer2023memgpt,kang2025memory}.
These systems typically process interaction sessions to extract and store factual knowledge or user preferences as manageable memory units, maintaining long-term continuity through mechanisms such as updating and forgetting~\cite{zhong2024memorybank}.
However, interrelated evidence is often fragmented across dispersed sessions, which hinders effective information utilization.

To mitigate this issue, recent studies have focused on associating related information distributed across different human-LLM interaction sessions.
These approaches~\cite{fang2025lightmem,xu2025mem} consolidate related knowledge to construct more effective and comprehensive memories.
Specifically, they achieve this by topic-based clustering~\cite{fang2025lightmem} or by transforming interaction sessions into structured memory fragments~\cite{xu2025mem}.
However, the resulting memory units are often fragmented or even contradictory, which disrupts the original logical and temporal relationships among facts.
Such memory construction prevents LLMs from fully leveraging the accumulated knowledge.

As illustrated in Figure~\ref{fig:intro}, for the given query, related memories may contain several inconsistent versions of factual knowledge regarding the same event, such as different estimates of a total travel distance of the ``Yellowstone trip'', including ``1200 miles'' and ``300 miles''. Due to the fragmented nature of these memory units, the model becomes confused, leading to incorrect answers. By introducing knowledge usage experiences, the model ignores misleading segments of the same journey, thereby enabling accurate reasoning. This observation highlights the crucial role of guiding LLMs to learn how to utilize memorized knowledge effectively, which aligns with the ``learning to learn'' principle in meta-learning~\cite{flavell1979metacognition}.

This paper proposes \method{}, a novel memory system framework augmented with a self-evolving meta-memory for memorized knowledge utilization.
During meta-memory evolution, \method{} first generates diverse reasoning responses conditioned on the user query, memory fragments, and the current meta-memory experiences. It then reflects on these responses under the guidance of correctness-based rewards to distill generalizable insights, and subsequently proposes updates to optimize the meta-memory.
After optimization, the meta-memory accumulates transferable knowledge utilization experiences, thereby guiding the LLM to more effectively and systematically exploit knowledge embedded in the memory fragments.

Our experimental results demonstrate the effectiveness of \method{} by consistently outperforming all baseline models with more than 3.6\% improvements. Notably, the advantages of \method{} generalize across different knowledge usage systems, highlighting its strong generalization capability.
Moreover, \method{} effectively benefits from out-of-domain data through constructed meta-memory and exhibits substantial scalability with respect to the amount of training data. This property indicates significant potential for enabling \method{} to continuously improve through long-term human-LLM interactions.

%% file: figure/intro.tex
\begin{figure}[t]
\centering
\includegraphics[width=\linewidth]{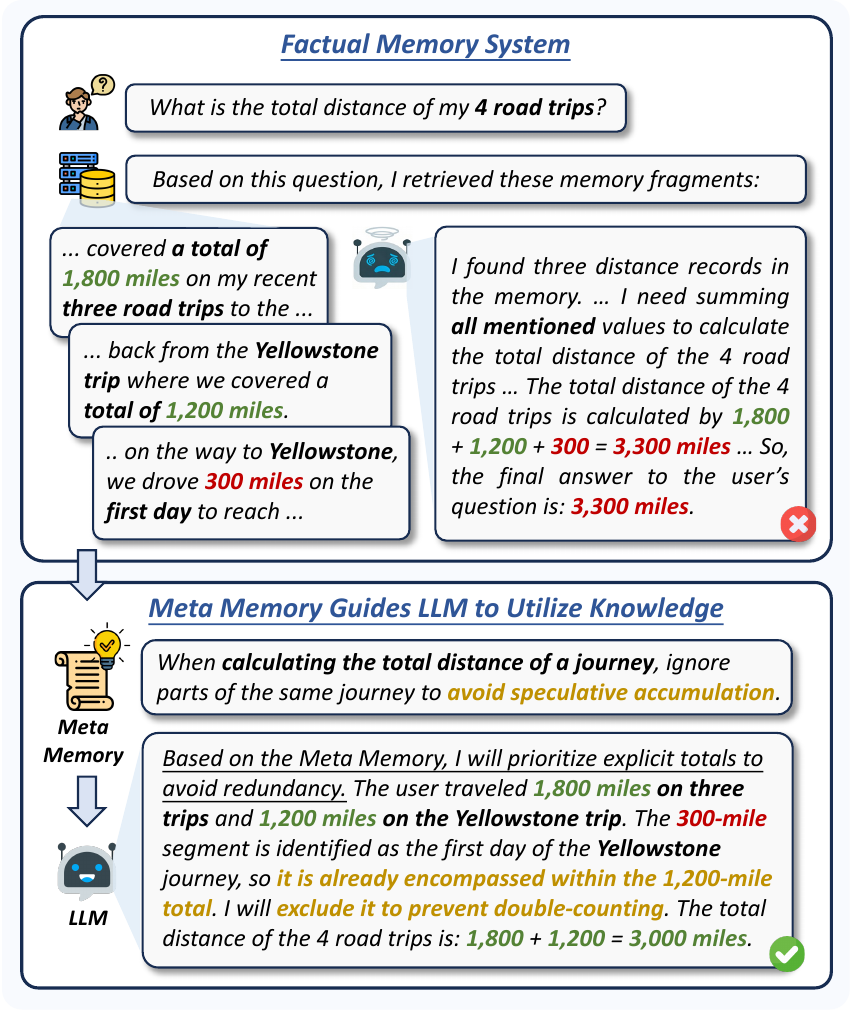}
\caption{Illustration of Our \method{} Framework. \textcolor[RGB]{84,130,53}{\textbf{Green}} indicates correct evidence, while \textcolor[RGB]{192,0,0}{\textbf{red}} represents distracting evidence. \method{} guides the LLM to effectively utilize knowledge from scattered memory fragments, thereby generating the correct answer.}
\label{fig:intro}
\end{figure}

%% file: section/2_related.tex
\section{Related Work}
Large Language Models (LLMs)~\cite{liu2024deepseek,yang2025qwen3,team2025kimi} have demonstrated remarkable capabilities across a wide range of complex tasks~\cite{he2021efficient,trivedi2023interleaving,liu5459034knowledge}. Nevertheless, due to the limited context window of LLMs and the ``lost-in-the-middle'' phenomenon~\cite{liu2024lost}, they face substantial difficulties when reasoning over long-horizon human--LLM interaction sessions. To mitigate these issues, existing approaches typically segment interaction histories into discrete chunks and retrieve relevant fragments to serve as model input~\cite{zhou2025llm,lewis2020retrieval}. However, such strategies neglect the inherent coherence of interaction sessions, often leading to the retrieval of fragmented contexts that omit critical information necessary for accurately answering the query.

Recent studies have focused on developing external memory systems to store long-term memory, aiming to better manage and utilize historical interaction sessions~\cite{packer2023memgpt,park2023generative,zhong2024memorybank}. These methods typically treat historical interaction sessions as a manageable memory resource and systematically maintain long-term memory through hierarchical storage mechanisms~\cite{kang2025memory} or continuous memory updating and forgetting~\cite{zhong2024memorybank}. Such designs enable LLMs to access more useful information across sessions and support long-horizon interactions. However, these methods lack an effective memory construction mechanism for systematically integrating information on similar topics that is scattered across different sessions, resulting in fragmented historical interaction memories.

To address this challenge, several studies have begun exploring effective memory construction methods, with a particular focus on extracting key information from the long context of human--LLM interactions. LightMem~\cite{fang2025lightmem} constructs long-term memory from historical interactions through a memory construction pipeline that integrates text compression, similar-topic aggregation, and summarization to capture the most crucial information. G-Memory~\cite{zhang2025g} organizes historical interactions into graph structures to capture relationships across different topical sessions, enabling the model to better access key information. While these memory systems effectively construct memory for LLMs, they often disrupt the original logical and temporal relationships among memory entries, preventing LLMs from fully exploiting the constructed memories. In contrast, \method{} facilitates memory utilization by maintaining a meta-memory that conditions and coordinates different retrieved memories.

%% file: section/3_methodology.tex
\input{figure/pipeline}
\section{Methodology}
In this section, we introduce \method{}, a memory system framework augmented by a self-evolving meta-memory. We first introduce the preliminaries of meta-memory augmented memory systems (Sec.~\ref{sec:preliminary}). Then, we detail the iterative optimizing process for meta-memory updation (Sec.~\ref{sec:exp}).

\subsection{The Architecture of Meta-Memory Augmented Memory System} \label{sec:preliminary}
To enable LLMs to effectively leverage long-horizon human-LLM interaction histories, existing methods~\cite{kang2025memory, fang2025lightmem} typically construct an explicit memory system. Such systems organize historical interactions into structured contextual memories by extracting and summarizing key information, thereby facilitating more accurate responses to a user query $q$.

Specifically, the memory system~\cite{fang2025lightmem} first identifies sessions that share identical or highly similar topics from long-horizon human-LLM interaction data. For each distinct topic, the corresponding sessions are aggregated into a generalized memory representation $m_i$, which captures the salient information shared across those topic-aligned sessions. Through this process, a memory set $\mathcal{M} = \{m_1, \dots, m_n\}$ covering diverse topics is constructed, where $n$ denotes the number of topics. Finally, the long-term memory set $\mathcal{M}$ is provided to the LLM as additional context to assist in answering the user question $q$:
\begin{equation}\small
y = \text{LLM}(\text{Instruct}_{\text{Gen}}(q,\mathcal{M})),
\end{equation}
where $\text{Instruct}_{\text{Gen}}$ denotes the instruction designed to prompt the LLM to generate the response $y$ conditioned on the long-term memory $\mathcal{M}$.
However, existing memory systems often disrupt the logical and temporal relationships within contextual memory during the chunking and summarization process. As a result, the memory unit $m$ in $\mathcal{M}$, although containing key evidence, may not be effectively identified or exploited by the LLM.

In contrast to prior methods~\cite{fang2025lightmem}, we propose \method{}, a meta-memory augmented system that enhances LLMs' ability to learn from and utilize external memory composed of key information. Specifically, \method{} learns a task-agnostic meta-memory $\mathcal{E}$ by training on a multi-task dataset $\mathcal{D}$, enabling the model to acquire transferable experience in memory utilization and to rapidly adapt to different tasks:
\begin{equation}\small
\mathcal{E}_T= \text{\method{}}(\mathcal{D},\text{LLM}(\cdot)),
\end{equation}
where the meta-memory $\mathcal{E}_T$ encapsulates learned experiences about how to effectively leverage memory across diverse tasks.
The training dataset is formulated as $\mathcal{D} = \{(q_i, a_i, \mathcal{M}_i)\}_{i=1}^N$, where $q_i$ denotes the question, $a_i$ is the ground-truth answer, and $\mathcal{M}_i$ is the corresponding memory set for $q_i$. During inference, the optimized meta-memory $\mathcal{E}_T$ is employed to guide the LLM in more effectively utilizing the external memory set $\mathcal{M}$ to generate the response $y$:
\begin{equation}\small
y = \text{LLM}(\text{Instruct}_{\text{Gen}}(q,\mathcal{M},\mathcal{E}_T)),
\end{equation}
where $\text{Instruct}_{\text{Gen}}$ denotes the instruction template used to prompt the LLM to answer the question.

\subsection{Meta-Memory Evolution for Learning to Use Memorized Contents} \label{sec:exp}
In this subsection, we describe the procedure for constructing the meta-memory with LLMs using the training dataset $\mathcal{D}$. We first present the overall workflow of meta-memory construction and then elaborate on the meta-memory update strategies based on self-reflection.

\textbf{The Workflow of Meta-Memory Learning.}
To balance efficiency and effectiveness, \method{} adopts a mini-batch learning strategy that partitions the training dataset $\mathcal{D}$ into multiple batches $\{\Tilde{\mathcal{D}}_1, \dots, \Tilde{\mathcal{D}}_T\}$. During optimization, we initialize the meta-memory as an empty state $\mathcal{E}_0 = \emptyset$ and iteratively update it across batches. This iterative process continues until all batches are processed, yielding the final meta-memory $\mathcal{E}_T$.

At each optimization step $t$, the current meta-memory $\mathcal{E}_t = \{e_t^1, \dots, e_t^n\}$ consists of multiple memory usage experiences, where each $e_t^i$ denotes an individual meta-memory unit. We utilize the $t$-th data batch $\Tilde{\mathcal{D}}_t$ to evolve $\mathcal{E}_t$ into an updated state $\mathcal{E}_{t+1}$. Formally, for each instance $(q, a, \mathcal{M})$ in the batch $\Tilde{\mathcal{D}}_t$, we employ a self-reflection-based optimization function, denoted as \texttt{Update}, to generate a candidate meta-memory update operation $\tilde{O}_t$ based on the current meta-memory state $\mathcal{E}_t$ and the batch data at step $t$:
\begin{equation}\small\label{eq:update}
\tilde{O}_t = \texttt{Update}(\Tilde{\mathcal{D}}_t, \mathcal{E}_t),
\end{equation}
where $\tilde{O}_t = \{o^1_t, \dots, o^m_t\}$ is a set of $m$ update operations for refining the meta-memory $\mathcal{E}_t$ according to the performance of the meta-memory-augmented memory system. Each operation $o_t^i$ belongs to one of the following three action types:
\begin{itemize}
\item \textit{Addition}: $o_t^i$ contains the identifier \texttt{ADD} that aims to append a newly generated meta-memory unit to $\mathcal{E}_t$.
\item \textit{Deletion}: $o_t^i$ contains the identifier \texttt{DEL} and specifies the index $j$ of an existing meta-memory unit $e_t^j \in \mathcal{E}_t$ to be removed. This operation is designed to eliminate obsolete or erroneous experience units.
\item \textit{Modification}: $o_t^i$ includes the identifier \texttt{MOD} and the index $j$ of the meta-memory unit $e_t^j \in \mathcal{E}_t$ to be updated, and replaces $e_t^j$ with a modified unit $e_{t+1}^j$.
\end{itemize}

Finally, we apply the update operations $\tilde{O}_t$ to the current meta-memory $\mathcal{E}_t$ using the execution function \texttt{Exec}, producing the optimized meta-memory $\mathcal{E}_{t+1}$:
\begin{equation}\small
\mathcal{E}_{t+1} = \texttt{Exec}(\tilde{O}_t,\mathcal{E}_t).
\end{equation}
This update process repeats until all $T$ batches are processed, resulting in the final meta-memory $\mathcal{E}_T$.

\textbf{Meta-Memory Update via Self-Reflection.}
We next detail how \method{} performs self-reflection over the current meta-memory state to derive optimization actions, corresponding to Eq.~\ref{eq:update}.

For each instance $(q, a, \mathcal{M})$ in the batch $\Tilde{\mathcal{D}}_t$ at optimization step $t$, \method{} first prompts the LLM to randomly sample a set of responses $\mathcal{T} = \{\tau_1, \dots, \tau_k\}$ conditioned on the current meta-memory state $\mathcal{E}_t$:
\begin{equation}\small
\tau_i \sim \text{LLM}(\text{Instruct}_\text{Gen}(q, \mathcal{M},\mathcal{E}_t)).
\end{equation}
We then employ a stronger model as a judge to evaluate each sampled response $\tau_i \in \mathcal{T}$ against the ground-truth answer $a$:
\begin{equation}\small
s_i=\texttt{Judge}(q, a, \mathcal{M},\tau_i),
\end{equation}
where $s_i$ is a binary indicator. Specifically, $s_i = 1$ if $\tau_i$ matches the ground truth $a$, and $s_i = 0$ otherwise, indicating that $\tau_i$ is inconsistent with $a$. Based on the evaluation score $s_i$, we prompt the LLM to conduct self-reflection on each response $\tau_i$, producing a reflection output $r_i$:
\begin{equation}\small
r_i=\text{LLM}(\text{Instruct}_\text{Reflect}(q, a, \mathcal{M},s_i,\tau_i)),
\end{equation}
where $\text{Instruct}_\text{Reflect}$ guides the LLM to analyze the sampled response $\tau_i$. Each reflection $r_i$ includes an assessment of the LLM-generated responses, explaining the reasoning and underlying causes of the correctness or incorrectness of $\tau_i$. We aggregate these reflections into a set $R = \{r_1, \dots, r_k\}$, which captures various insights over the sampled responses.
We treat the reflection set $R$ as feedback signals and leverage the instruction $\text{Instruct}_\text{Action}$ to prompt the LLM to generate an edit action $o$ for the current meta-memory $\mathcal{E}_t$ at step $t$:
\begin{equation}\small \label{eq:action_gen}
o = \text{LLM}(\text{Instruct}_{\text{Action}}(q,a,R,\mathcal{E}_t)).
\end{equation}
We aggregate the actions $o$ generated from all instances $(q,a,R)$ in $\Tilde{\mathcal{D}}_t$ to form an action set $O_t$.
Since actions proposed by different instances may conflict, we further apply a filtering instruction, $\text{Instruct}_\text{Filter}$, to prompt the LLM to resolve such conflicts and produce a consistent action set $\tilde{O}_t$ for meta-memory update:
\begin{equation}\small
\tilde{O}_t = \text{LLM}(\text{Instruct}_{\text{Filter}}(O_t,\mathcal{E}_t)).
\end{equation}

%% file: figure/pipeline.tex
\begin{figure*}[t]
\centering
\includegraphics[width=\linewidth]{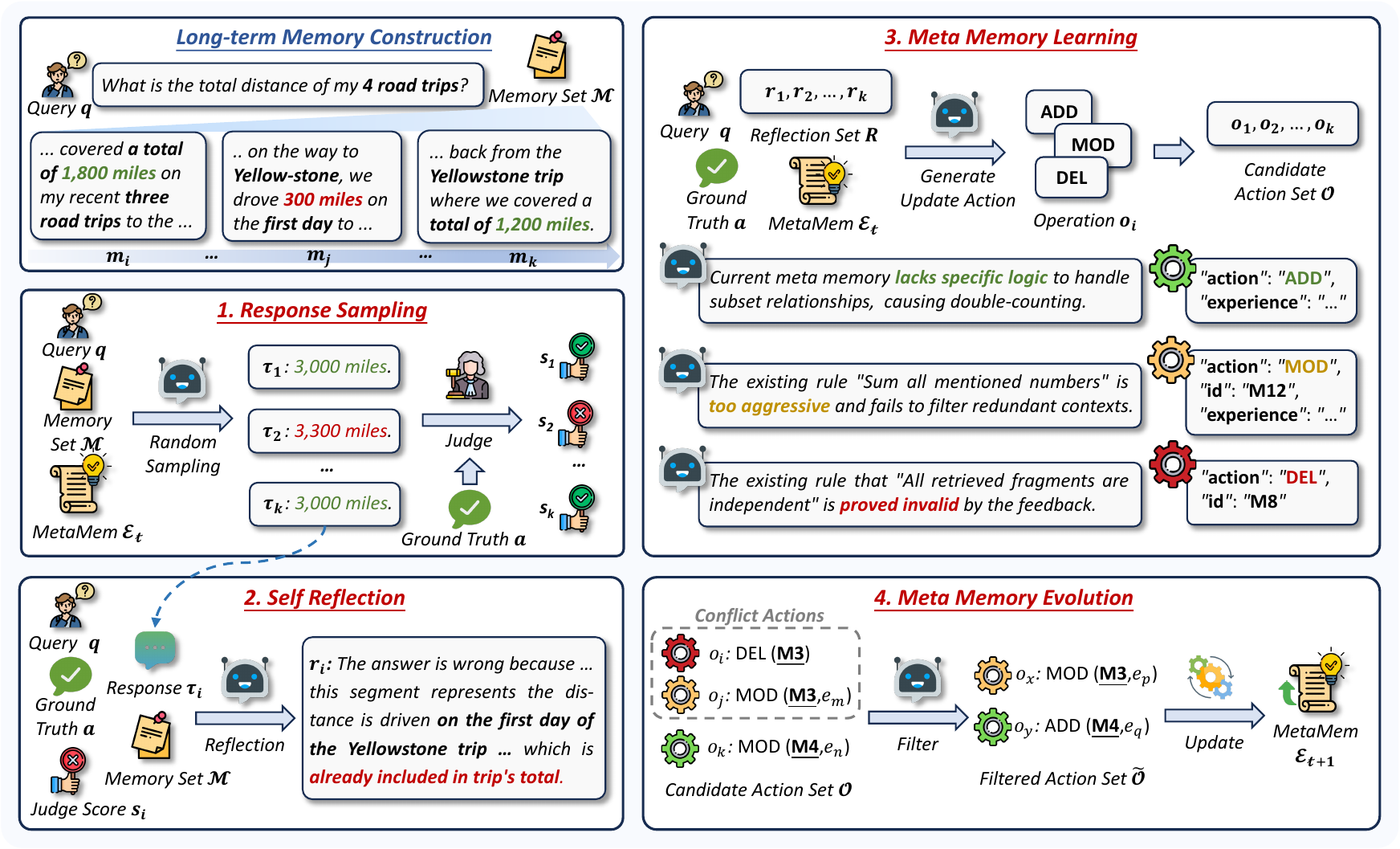}
\caption{Overview of \method{} Model. \method{} evolves through environmental feedback, guiding the memory system to utilize factual knowledge through meta-memory.}
\label{fig:pipeline}
\end{figure*}

%% file: section/4_experiment.tex
\section{Experimental Methodology}
This section describes the datasets, evaluation metrics, and baseline methods. Then, we introduce the implementation details of our experiments.

\textbf{Datasets.} 
We evaluate \method{} on LongMemEval~\cite{wu2024longmemeval}, a benchmark tailored for assessing long-horizon human-LLM interactions within multi-task dialogue scenarios.
We additionally use the ShareGPT~\cite{ShareGPT-Chinese-English-90k} dataset as an out-of-domain corpus for \method{} construction for a generalization test.
More details are provided in Appendix~\ref{app:dataset}. Additionally, since LongMemEval contains only 500 test samples, we employ 5-fold cross-validation for evaluation, where each fold allocates 400 samples for Meta-Memory construction and 100 samples for testing. More details are shown in Appendix~\ref{app:fold}.

\textbf{Evaluation Metrics.} 
Following \citet{fang2025lightmem}, we adopt accuracy as the evaluation metric, which is defined as the proportion of correctly answered questions.
We evaluate \method{} on the LongMemEval benchmark across four dimensions.
Single-Session tasks (Single User, Single Assistant, and Single Preference) assess intra-session retention. 
Multi-Session tasks evaluate information integration across disjoint sessions. 
Temporal Reasoning tasks test the modeling of chronological dependencies. 
Knowledge Update tasks evaluate the capability of models to adapt to newly introduced conflicting information.

\textbf{Baselines.}
We compare \method{} with several representative baseline methods.
The Full Text model directly feeds the entire dialogue history into the model as context for question answering.
We further include RAG~\cite{lewis2020retrieval} and MapReduce~\cite{zhou2025llm} as baselines; both aim to identify and aggregate query-relevant chunks to produce the final response.
Moreover, we compare against several memory-based systems that maintain persistent storage to track factual updates over long-term interactions, including Mem0~\cite{chhikara2025mem0}, A-Mem~\cite{xu2025mem}, MemoryOS~\cite{kang2025memory}, and LightMem~\cite{fang2025lightmem}.
Specifically, Mem0 and A-Mem manage long-term interactions by maintaining persistent vector databases that automatically update user profiles and factual records.
MemoryOS formulates memory management as an operating system, performing hierarchical read and write operations to ensure memory consistency and coherence.
LightMem serves as our primary baseline; it optimizes knowledge storage by incorporating a text compression module, topic-aware short-term memory, and long-term memory maintenance mechanisms.
For a fair comparison, all methods employ the same backbone LLM and retriever in our experiments.

\input{table/overall}
\textbf{Implementation Details.} 
We conduct experiments using Qwen3-30B-A3B-Instruct~\cite{yang2025qwen3} and Llama3.1-70B-Instruct~\cite{grattafiori2024llama} as backbone models, and employ Qwen3-235B-A22B~\cite{yang2025qwen3} for response judgment. 
The prompt templates used in \method{} are provided in Appendix~\ref{app:prompt}.
We leverage LightMem~\cite{fang2025lightmem} for factual memory ($\mathcal{M}$) construction, and utilize all-MiniLM-L6-v2~\cite{wang2020minilm} as the embedding model for memory retrieval.
Further implementation details are provided in Appendix~\ref{app:exp}.

%% file: table/overall.tex
\begin{table*}[t]
\centering
\small
\begin{tabular}{lcccccccc}
\toprule
\multirow{2}{*}{\textbf{Method}} & \textbf{Single} & \textbf{Single} & \textbf{Multi} & \textbf{Temporal} & \textbf{Knowledge} & \textbf{Single} & \multirow{2}{*}{\textbf{Avg.}} \\
& \textbf{User} & \textbf{Assistant} & \textbf{Session} & \textbf{Reasoning} & \textbf{Update} & \textbf{Preference} & \\
\midrule
\rowcolor{gray!8}\multicolumn{8}{l}{\textbf{\textit{Qwen3-30B-A3B-Instruct}}} \\
\midrule
Full Text & 81.13 & \textbf{88.37} & 34.29 & 27.52 & 66.67 & 62.50 & 51.50 \\
RAG & 86.79 & \textbf{88.37} & 46.67 & 49.54 & 66.67 & 75.00 & 62.25 \\
MapReduce~\citeyearpar{zhou2025llm} & 81.13 & 81.40 & 38.46 & 36.45 & 72.31 & 62.50 & 55.30 \\
Mem0~\citeyearpar{chhikara2025mem0} & 82.86 & 87.50 & 36.09 & 33.08 & \underline{76.92} & 50.00 & 54.80 \\
A-Mem~\citeyearpar{xu2025mem} & 85.85 & 86.05 & 52.38 & 51.38 & 72.73 & 41.67 & 62.88 \\
MemoryOS~\citeyearpar{kang2025memory} & 69.81 & 79.07 & 37.14 & 28.44 & 59.09 & 29.17 & 46.75 \\
LightMem~\citeyearpar{fang2025lightmem} & \underline{90.57} & 32.56 & \underline{62.86} & \underline{64.22} & 75.76 & \underline{91.67} & \underline{67.50} \\
\method{} & \textbf{90.70} & 38.14 & \textbf{69.24} & \textbf{69.60} & \textbf{79.18} & \textbf{94.16} & \textbf{71.90} \\
\midrule
\rowcolor{gray!8}\multicolumn{8}{l}{\textbf{\textit{Llama3.1-70B-Instruct}}} \\
\midrule
Full Text & 58.49 & 72.09 & 21.90 & 22.02 & 57.58 & 33.33 & 38.75 \\
RAG & 84.91 & \textbf{90.70} & 48.57 & 47.71 & 62.12 & 58.33 & 60.50 \\
MapReduce~\citeyearpar{zhou2025llm} & 86.79 & 83.72 & 15.24 & 26.61 & 71.21 & 62.50 & 47.25 \\
Mem0~\citeyearpar{chhikara2025mem0} & 81.43 & \underline{89.29} & 34.15 & 35.88 & \underline{77.18} & 38.33 & 54.62 \\
A-Mem~\citeyearpar{xu2025mem} & 84.29 & 88.64 & 56.72 & \underline{57.45} & 73.56 & 32.50 & 62.88 \\
MemoryOS~\citeyearpar{kang2025memory} & \textbf{90.57} & 83.72 & 51.43 & 55.05 & 62.12 & 50.00 & 62.75 \\
LightMem~\citeyearpar{fang2025lightmem} & \underline{90.36} & 22.36 & \underline{75.31} & 52.52 & 76.92 & \textbf{70.24} & \underline{66.17} \\
\method{} & 90.28 & 25.64 & \textbf{75.50} & \textbf{66.28} & \textbf{77.76} & \underline{66.12} & \textbf{69.08} \\
\bottomrule
\end{tabular}
\caption{Overall Performance on the LongMemEval~\cite{wu2024longmemeval} Benchmark. The best results are marked in \textbf{bold}, while the second-best results are \underline{underlined}.} \label{tab:overall}
\end{table*}

%% file: section/5_results.tex
\section{Experimental Results}
In this section, we show the overall performance of \method{} and conduct ablation studies to investigate the contribution of each component. We then analyze the evolution of meta-memory units during the self-evolution process. Finally, case studies are provided in Appendix~\ref{app:case}.

\subsection{Overall Performance}
In this section, we first report the overall performance of \method{} across two backbone models, followed by an assessment of its generalization capability in the out-of-domain setting.

Table~\ref{tab:overall} presents the performance of \method{} across the Qwen3-30B-A3B-Instruct and Llama3.1-70B-Instruct backbone models.
Overall, \method{} consistently achieves state-of-the-art performance, outperforming the strongest baseline by over 3.6\%. Notably, \method{} exhibits superior robustness on Multi-Session and Temporal Reasoning tasks, substantially outperforming our primary baseline LightMem~\cite{fang2025lightmem}. This result further confirms the effectiveness of the augmented meta-memory module.

Compared to the Full Text baseline, chunking-based approaches such as RAG and MapReduce exhibit moderate improvements on Multi-Session and Temporal Reasoning tasks. However, their reliance on finding more query-related evidence leads to suboptimal performance in conducting static knowledge and continuous reasoning.
Stateful memory systems, including Mem0, A-Mem, and MemoryOS, further improve performance on Single-User and Knowledge Update tasks by maintaining persistent knowledge across sessions.
However, these methods update memory content in a largely static manner and are not tailored to the LLMs during the reasoning process, resulting in limited gains on complex long-horizon reasoning tasks.
In contrast, \method{} leverages a self-evolving meta-memory that is dynamically updated through feedback while handling different tasks, enabling LLMs to effectively utilize static knowledge stored in existing memories and produce more accurate responses.

\input{figure/general}
\input{table/ablation}
Additionally, we train \method{} separately on out-of-domain data (ShareGPT) and in-domain data (LongMemEval), resulting in two variant models, \method{} (ShareGPT) and \method{} (LongMemEval), to assess the cross-domain generalization capability of meta-memory construction across different datasets.
As shown in Figure~\ref{fig:general_domain}, both \method{} (ShareGPT) and \method{} (LongMemEval) consistently outperform the baseline LightMem model, regardless of whether the training data is in-domain or out-of-domain. These results indicate that the meta-memory learning process of \method{} is not dependent on domain-specific data and exhibits strong generalization ability, effectively enhancing the LLM's capacity to leverage external memory. Moreover, \method{} (LongMemEval), trained on in-domain data, achieves the best performance due to the closer alignment between training and evaluation, highlighting the important role of a tailored meta-memory in effective memory utilization.

\subsection{Ablation Study}
To investigate the contribution of each component in \method{}, we conduct an ablation study on the Qwen3-30B-A3B-Instruct and Llama3.1-70B-Instruct backbone models.

As shown in Table~\ref{tab:ablation}, we first evaluate two variants: \method{} w/o Evolution, which disables the evolution of the meta-memory, and \method{} w/o Self-Reflection, which removes the intermediate self-reflection step and directly generates optimization actions from the original responses.
The evaluation results show that, compared with \method{} w/o Evolution, \method{} achieves more than 3\% performance improvements, demonstrating the critical role of the iterative optimization mechanism in constructing a more tailored meta-memory to guide LLMs in effectively utilizing knowledge.
These results suggest that meta-memory should be continuously updated based on feedback from dynamic environments, rather than remaining static.
By learning across diverse tasks, the meta-memory is iteratively updated to capture generalizable knowledge and experience, enabling LLMs to handle different tasks more effectively instead of overfitting to task-specific factual knowledge.
When removing the self-reflection step (\method{} w/o Self-Reflection), the performance of \method{} usually degrades, highlighting the effectiveness of the self-reflection module.
This observation indicates that the self-reflection process complements the meta-memory update mechanism by functioning as a critic, producing high-quality insights through contrastive comparison between generated responses and ground-truth answers.

Furthermore, to assess the generalizability of \method{} across different knowledge utilization systems, we apply \method{} to two distinct methods, namely Full Text and RAG, resulting in two variants: Full Text (w/ \method{}) and RAG (w/ \method{}).
Experimental results demonstrate that both variants achieve consistent performance improvements over vanilla memory models, showing their effectiveness and generalization capability.
These results confirm that the effectiveness of \method{} is decoupled from the underlying memory system, showing that \method{} can robustly adapt to and enhance diverse memory systems without relying on system-specific characteristics.

\subsection{Characteristics of Meta-Memory Evolution during Optimization}
As shown in Figure~\ref{fig:training}, we analyze how memory usage experiences stored in the meta-memory evolve as training progresses, and examine the corresponding performance changes induced by the evolved meta-memory. This experiment adopts Qwen3-30B-A3B-Instruct as the backbone model.

As shown in Figure~\ref{fig:training:a}, we first investigate the evolution of meta-memory units throughout the training process. Specifically, we employ GLM-4.5~\cite{zeng2025glm} to classify memory usage experience units extracted from meta-memory checkpoints at each training step into two categories: Specific units, which capture task-specific memory usage experiences, and General units, which represent task-agnostic and broadly applicable experiences. Detailed classification instructions are provided in Appendix~\ref{app:prompt}. The results demonstrate that, as training proceeds, the proportion of General units increases steadily from approximately 65\% to over 80\%, and eventually stabilizes at this higher level. This observation indicates that the optimization process progressively shifts the meta-memory from encoding task-specific experiences toward consolidating more generalizable memory usage patterns, thereby enabling the memory system to transfer and apply learned experiences more effectively across diverse tasks.
\input{figure/training}

Additionally, as shown in Figure~\ref{fig:training:b}, we report the task-wise accuracy of \method{} under different stages of meta-memory evolution. The results show that during the early training stages, \method{} exhibits consistent performance improvements across all tasks, suggesting that updating meta-memory using more general memory usage experiences effectively enhances the system's ability to leverage external memory. However, except for the Knowledge Update task, performance on most tasks begins to decline after peaking at step~5. This trend suggests that excessive training may cause the meta-memory to overfit specific knowledge update behaviors learned from later optimization, thereby degrading the performance of \method{}.

\input{figure/scaling}
\subsection{Effectiveness of \method{} with Human-LLM Interaction Scaling}
To comprehensively evaluate the effectiveness of \method{} under human-LLM interaction scaling, we report both task accuracy and predictive confidence in Figure~\ref{fig:scaling}. All models are implemented based on Qwen3-30B-A3B-Instruct and optimized using ShareGPT data for meta-memory training.

As illustrated in Figure~\ref{fig:scaling:a}, under extremely limited data regimes, \method{} slightly underperforms the baseline.
This initial degradation is likely due to the instability of the learned meta-memory during the early stage of optimization.
Nevertheless, \method{} quickly adapts and surpasses the baseline with only 400 training samples.
As the data scale continues to increase, \method{} maintains a stable upward performance trend, eventually widening the performance gap to 2.9\% at 1,000 samples.
These results indicate that \method{} is highly data-efficient, demonstrating a strong ability to learn how to effectively facilitate knowledge utilization within the model as training data scales.

Furthermore, Figure~\ref{fig:scaling:b} presents the evolution of perplexity.
Overall, we observe a monotonic reduction in PPL from 3.59 to 3.19, suggesting that increased data exposure consistently reduces predictive uncertainty.
Notably, the rate of improvement exhibits diminishing marginal returns: the PPL decreases substantially when scaling the data from 200 to 800 samples, but largely plateaus between 800 and 1,000 samples.
This observation suggests that the core meta-memory patterns can be effectively acquired with a moderate amount of data, enabling LLMs to better leverage memorized knowledge and generate more confident responses.

%% file: figure/general.tex
\begin{figure}[t]
\centering
\includegraphics[width=0.9\linewidth]{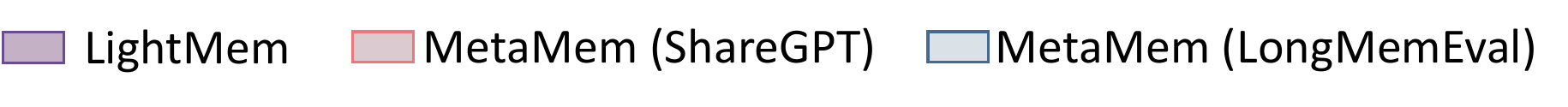}
\subfigure[Performance on Qwen3-30B-A3B-Instruct Model.]{
\includegraphics[width=0.46\linewidth]{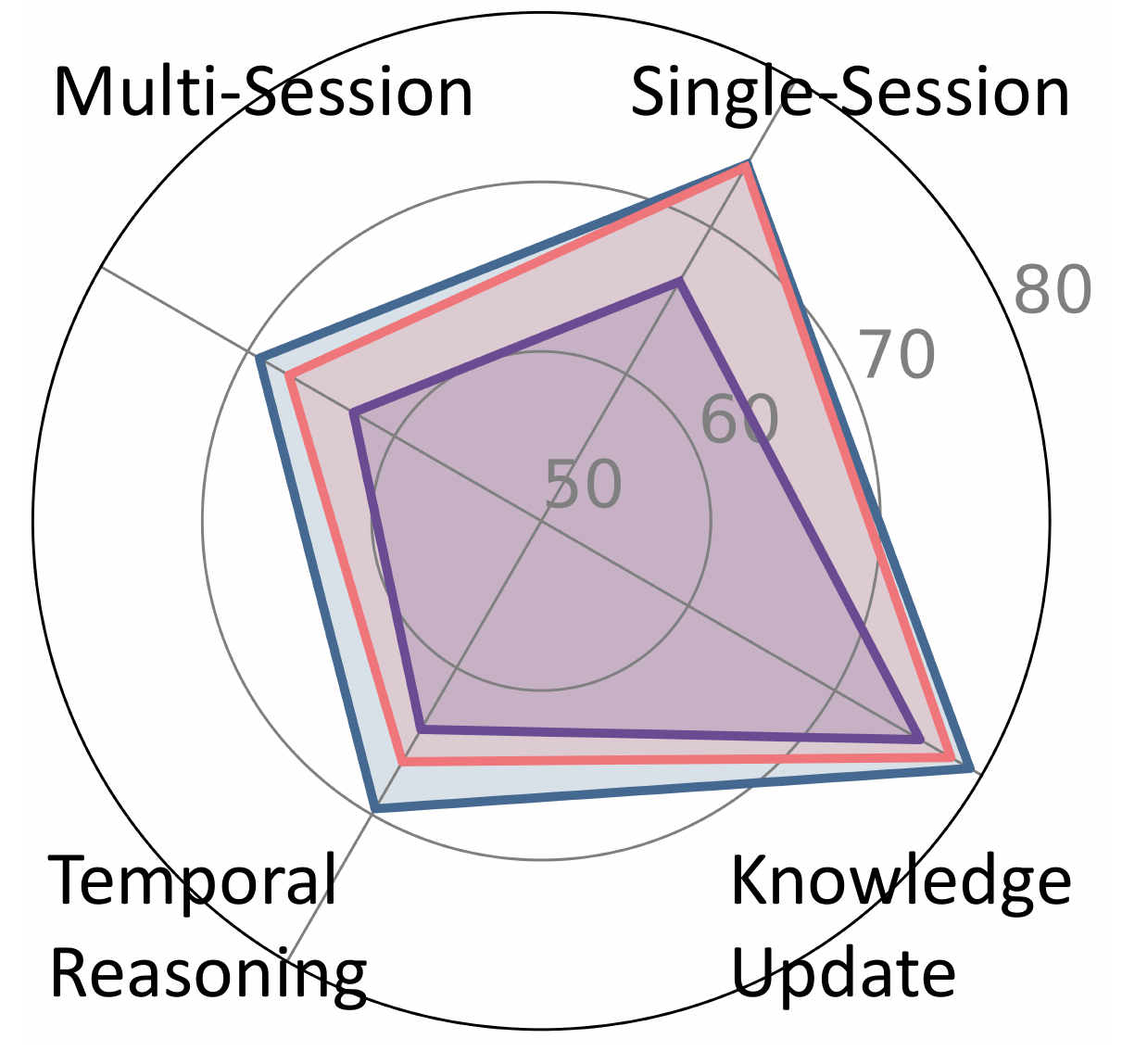}
\label{fig:scaling:a}
}
\subfigure[Performance on Llama3.1-70B-Instruct Model.]{
\includegraphics[width=0.46\linewidth]{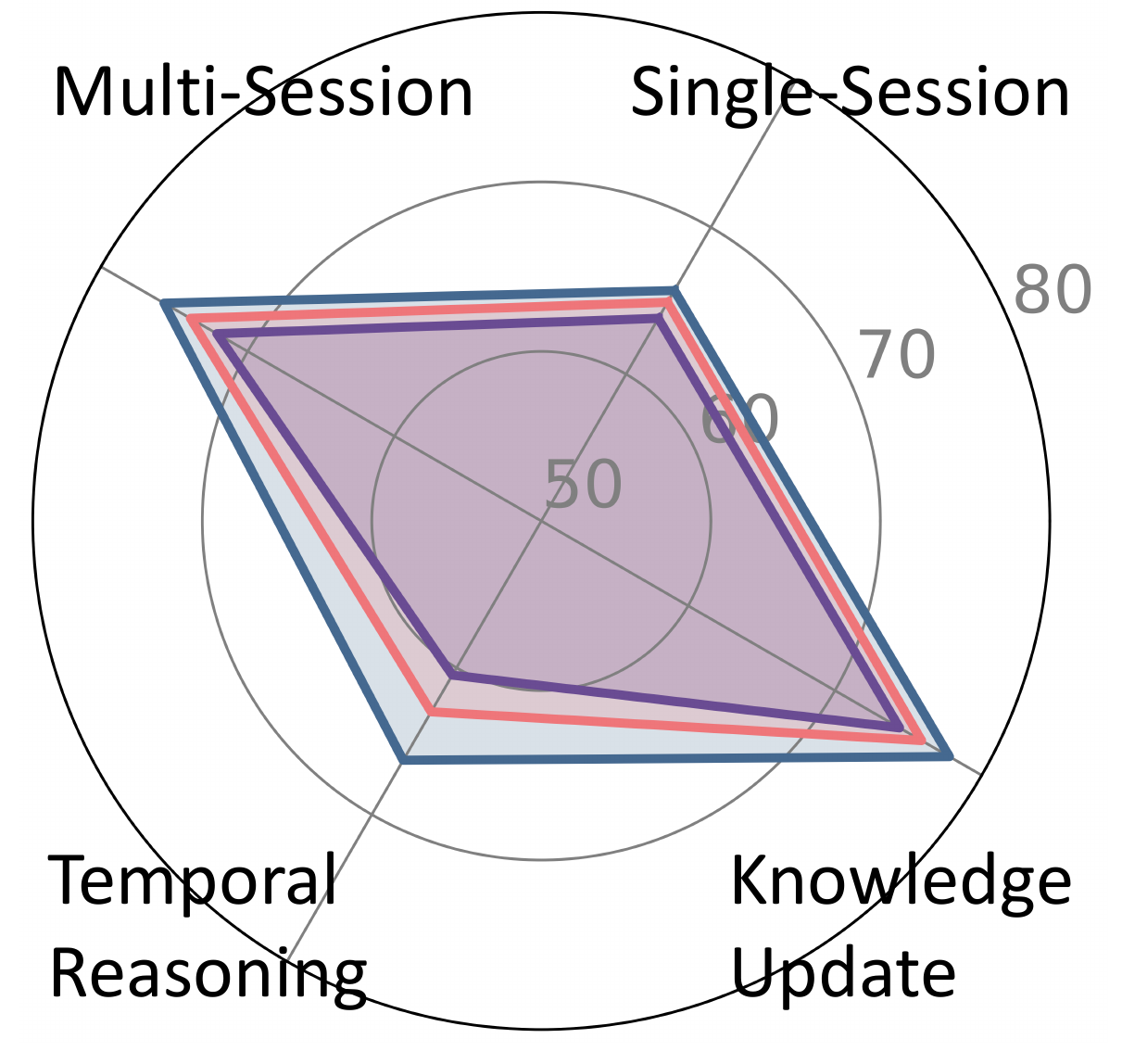}
\label{fig:scaling:b}
}
\caption{Generalization Ability of \method{}. Performance is reported under Qwen3-30B-A3B-Instruct and Llama3.1-70B-Instruct backbone models.}
\label{fig:general_domain}
\end{figure}

%% file: table/ablation.tex
\begin{table*}[t]
\centering
\small
\begin{tabular}{lcccccccc}
\toprule
\multirow{2}{*}{\textbf{Method}} & \textbf{Single} & \textbf{Single} & \textbf{Multi} & \textbf{Temporal} & \textbf{Knowledge} & \textbf{Single} & \multirow{2}{*}{\textbf{Avg.}} \\
& \textbf{User} & \textbf{Assistant} & \textbf{Session} & \textbf{Reasoning} & \textbf{Update} & \textbf{Preference} & \\
\midrule
\rowcolor{gray!8}\multicolumn{8}{l}{\textbf{\textit{Qwen3-30B-A3B-Instruct}}} \\
\midrule
\rowcolor{blue!8}\method{} & 90.70 & 38.14 & 69.24 & 69.60 & 79.18 & 94.16 & 71.90 \\
w/o Self-Reflection & 90.57 & 34.35 & 66.24 & 68.20 & 78.19 & 91.67 & 70.02 \\
w/o Evolution & 90.57 & 32.56 & 64.71 & 66.67 & 76.92 & 91.67 & 68.33 \\
\hdashline
Full Text & 81.13 & 88.37 & 34.29 & 27.52 & 66.67 & 62.50 & 51.50 \\
\rowcolor{blue!8}w/ \method{} & 86.79 & 88.37 & 38.10 & 25.69 & 69.70 & 62.50 & 53.25 \\
\hdashline
RAG & 86.79 & 88.37 & 46.67 & 49.54 & 66.67 & 75.00 & 62.25 \\
\rowcolor{blue!8}w/ \method{} & 90.57 & 86.05 & 51.43 & 54.13 & 66.67 & 91.67 & 66.42 \\
\midrule
\rowcolor{gray!8}\multicolumn{8}{l}{\textbf{\textit{Llama3.1-70B-Instruct}}} \\
\midrule
\rowcolor{blue!8}\method{} & 90.28 & 25.64 & 75.50 & 66.28 & 77.76 & 66.12 & 69.08 \\
w/o Self-Reflection & 90.28 & 25.64 & 75.42 & 64.68 & 77.34 & 69.45 & 68.33 \\
w/o Evolution & 90.11 & 23.88 & 75.42 & 63.15 & 77.34 & 69.45 & 66.42 \\
\hdashline
Full Text & 58.49 & 72.09 & 21.90 & 22.02 & 57.58 & 33.33 & 38.75 \\
\rowcolor{blue!8}w/ \method{} & 59.84 & 72.09 & 23.15 & 23.88 & 59.21 & 36.67 & 40.42 \\
\hdashline
RAG & 84.91 & 90.70 & 48.57 & 47.71 & 62.12 & 58.33 & 60.50 \\
\rowcolor{blue!8}w/ \method{} & 86.79 & 88.37 & 51.43 & 49.54 & 69.70 & 66.12 & 66.42 \\
\bottomrule
\end{tabular}
\caption{Ablation Study of \method{}.} \label{tab:ablation}
\end{table*}

%% file: figure/training.tex
\begin{figure}[t]
\centering
\subfigure[Unit Categories of Meta-Memory during Training.]{
\includegraphics[width=0.46\linewidth]{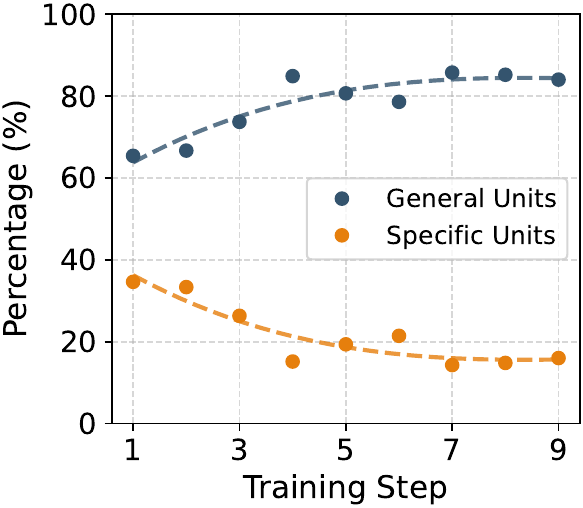}
\label{fig:training:a}
}
\hfill
\subfigure[Accuracy Trends across Different Sub-Tasks.]{
\includegraphics[width=0.46\linewidth]{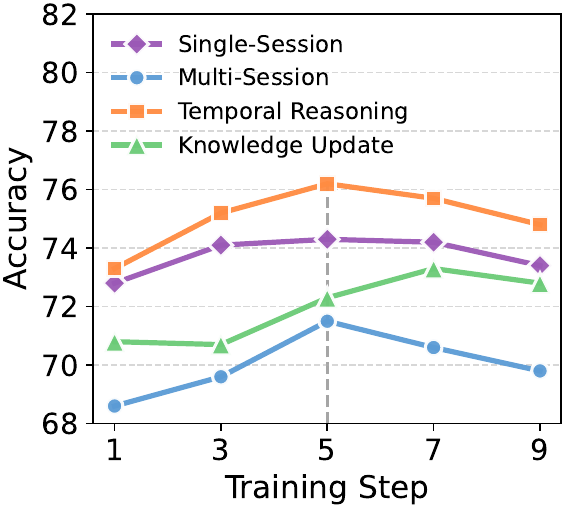}
\label{fig:training:b}
}
\caption{Performance of the Evolved Meta-Memory across Different Training Steps.}
\label{fig:training}
\end{figure}

%% file: figure/scaling.tex
\begin{figure}[t]
\centering
\subfigure[QA Accuracy.]{
\includegraphics[width=0.46\linewidth]{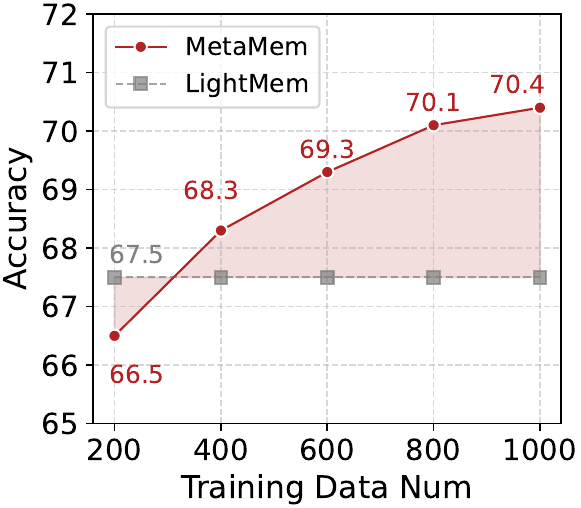}
\label{fig:scaling:a}
}
\subfigure[Predictive Uncertainty.]{
\includegraphics[width=0.46\linewidth]{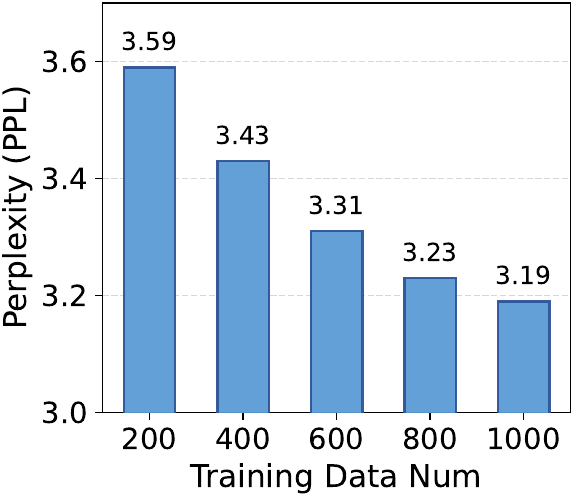}
\label{fig:scaling:b}
}
\caption{Effectiveness of \method{} Optimized with Different Data Scales. Figure~\ref{fig:scaling:a} illustrates the model performance as the training dataset size increases. Figure~\ref{fig:scaling:b} reports the perplexity of model outputs conditioned on the learned meta-memory.}
\label{fig:scaling}
\end{figure}

%% file: section/6_conclusion.tex
\section{Conclusion}
This paper proposes \method{}, a novel framework that equips LLM-based memory systems with a self-evolving meta-memory to effectively teach LLMs to learn to use memorized knowledge.
Extensive experimental results demonstrate that \method{} achieves state-of-the-art performance, while exhibiting strong cross-domain generalization and scalability, thereby benefiting long-horizon human-LLM interactions.

%% file: section/7_limitations.tex
\section*{Limitations}
Although our approach demonstrates convincing performance in guiding LLMs to utilize knowledge effectively, it has some limitations. 
First, our approach relies on an LLM-based judgment model to evaluate complex human-LLM interactions during meta-memory construction. 
Since real-world human-LLM interactions are highly nuanced and laden with diverse, implicit signals, establishing an accurate evaluation mechanism remains a persistent challenge.
Second, although meta-memory trained on out-of-domain data demonstrates strong generalization across diverse scenarios, our experiments show that it remains inferior to in-domain training. 
In practice, however, high-quality data for specific professional scenarios is often limited, making it difficult to fully optimize a scenario-specific meta-memory.
Consequently, we resort to a general-purpose meta-memory trained on large-scale heterogeneous data as a practical compromise, which prioritizes broad applicability.

\section*{Ethical Considerations}
This work does not raise significant ethical concerns. 
The experiments rely on the ShareGPT dataset, which is publicly available and has undergone strict anonymization and de-identification processes. We ensure that the data used in this study does not contain any personally identifiable information or sensitive private content. 
Consequently, our research involves no privacy risks regarding real-world users. 
Furthermore, our proposed method focuses on optimizing memory mechanisms and does not introduce additional risks regarding the generation of harmful or biased content. 
All experiments are conducted in compliance with standard ethical guidelines for academic research.

%% file: section/8_appendix.tex
\section{Appendix} \label{sec:appendix}
\FloatBarrier

\subsection{License} \label{app:license}
All datasets used in this work are released under licenses that permit academic research and non-commercial use, including LongMemEval (MIT License) and ShareGPT (Apache License 2.0).

\input{table/dataset}
\subsection{Additional Dataset Details} \label{app:dataset}
We employ both the LongMemEval~\cite{wu2024longmemeval} and the ShareGPT~\cite{ShareGPT-Chinese-English-90k} datasets for the construction and evaluation of \method{}. 
For LongMemEval, we adhere to the evaluation protocol established by \citet{fang2025lightmem}, with detailed dataset statistics presented in Table~\ref{tab:dataset}. 
For ShareGPT, which serves as an out-of-domain corpus for \method{} construction, we retain conversations that have at least 8 turns and randomly sample 1,000 instances with a random seed of 42. 
During inference, the LLM is provided with the first $n-1$ turns of a conversation and tasked with predicting the final user question.

\subsection{Additional Experimental Details} \label{app:exp}
For all the baselines, we utilize the same generation settings during memory construction. 
Specifically, we employ all-MiniLM-L6-v2~\cite{wang2020minilm} as the embedding model and set the retrieval top-$k$ to 20. 
For the RAG method, each dialogue turn is treated as a distinct chunk, and we concatenate the timestamp information to each chunk before embedding. 
During inference, we utilize the SGLang\footnote{\url{https://github.com/sgl-project/sglang}} framework to deploy the LLM server for high-performance inference. 
We standardize the LLM sampling parameters with a temperature of 0.0, a top-$p$ of 0.8, and a maximum token limit of 2000. 
During meta-memory construction, we maintain the sampling parameters at a fixed temperature of 0.7, a top-$p$ of 0.95, and a maximum output length of 4000 tokens, performing 5 samples per instance. 
The model is trained for 5 epochs with a batch size of 50.

\input{table/infer_time}
\subsection{Analysis of Inference Efficiency}
Table~\ref{tab:infer_time} reports the inference efficiency of \method{} across Qwen3-30B-A3B-Instruct and Llama3.1-70B-Instruct backbone models.
We implement \method{} across three distinct baselines--Full Text, RAG, and LightMem--to explicitly evaluate the inference latency overhead introduced by our approach.
For the Full Text method, \method{} notably accelerates inference by curbing verbosity and hallucinations, resulting in reduced token counts and latency. 
For the chunking-based method RAG and the stateful memory system LightMem, \method{} incurs negligible overhead due to the encoding of additional meta-directives. 
These results confirm that \method{} enhances response quality with minimal computational cost.

\input{table/k_fold} 
\subsection{Details of 5-Fold Cross-Validation} \label{app:fold}
We utilize random sampling to partition the dataset into five distinct subsets. 
To ensure reproducibility, we set the random sample seed to 42.
For each fold, a subset of 100 samples is reserved as the independent test set, while the remaining data is partitioned into 350 samples for meta-memory training and 50 samples for validation.
Table~\ref{tab:k_fold} shows the detailed results of our 5-fold cross-validation between LightMem~\cite{fang2025lightmem} and \method{}. 
We reported the final results by calculating the average performance across these five folds.

\input{figure/unit_cnt}
\subsection{Evolution of Meta-Memory Unit Count and Accuracy across Training Steps}
We further investigate the correlation between the count of the learned meta-memory units and task performance during the training process. 
Figure~\ref{fig:unit_count} illustrates the evolution of unit counts and accuracy on the Qwen3-30B-A3B-Instruct and Llama3.1-70B-Instruct backbone models.

As observed in both models, the number of meta-memory units exhibits a rapid growth phase in the early stages. 
This accumulation of units correlates positively with performance improvements, indicating that the model is effectively acquiring necessary reasoning directives. 
However, a turning point appears around step 5. While the unit count continues to rise or stabilize at a high level (approximately 25-27 units), the accuracy peaks at step 5 and subsequently exhibits a downward trend. 
This suggests that beyond this optimal point, the continued accumulation of rules may introduce redundancy or overly specific constraints, leading to overfitting and a slight degradation in general reasoning performance.

\subsection{Sensitivity Analysis of \method{} to Judge Model Choice} \label{app:judge}
\input{table/judge_sensitivity}

Since the meta-memory evolution process relies on LLM-based correctness feedback, an important question is whether the effectiveness of \method{} is sensitive to the particular judge model used during optimization. In the main experiments, we adopt Qwen3-235B-A22B as the default judge model for response evaluation. To assess the robustness of \method{} to the judge choice, we replace the default judge with two substantially different alternatives, namely Qwen3-30B-A3B-Instruct and Llama3.1-70B-Instruct, and rerun the meta-memory construction and evaluation pipeline under the same experimental setting.

Table~\ref{tab:judge_sensitivity} reports the overall accuracy on LongMemEval under different judge models. We observe that the performance of \method{} remains highly stable across judges, with only minor fluctuations in the final results. In all cases, \method{} consistently outperforms the underlying LightMem baseline by a clear margin.

\input{table/case_study}
\subsection{Case Study} \label{app:case}
In Table~\ref{tab:case_study}, we present a qualitative case study to illustrate how the learned meta-memory effectively guides the LLM to utilize knowledge from both aggregated facts and fragmented numerical evidence during long-horizon interaction sessions.

The user queries the total distance driven across four road trips. 
The retrieved memory fragments contain a mixture of numerical signals at different granularities, such as explicit summaries (\textit{e.g.}, ``1,800 miles across three trips'') and constituent details (\textit{e.g.}, ``300 miles on the first day''). 
The baseline model (LightMem~\cite{fang2025lightmem}) fails to comprehend the inclusion relationship between them. 
It treats the partial distance (300 miles) as an independent event rather than a subset of the total trip, leading to erroneous double-counting. 
Furthermore, it fails to filter out irrelevant planning-related estimates, resulting in a chaotic and incorrect summation.

In contrast, \method{} successfully leverages a learned meta-memory unit that instructs the model to prioritize verified cumulative statistics over fragmented partial data. 
Guided by this experience, the model correctly identifies that the ``300-mile'' segment is inherently encompassed within the ``1,200-mile Yellowstone trip'' summary. Consequently, \method{} filters out the redundant partial values and correctly composes the two high-level aggregates (1,800 miles + 1,200 miles), yielding the precise total. 
This case demonstrates that the learned meta-memory equips the LLM with the ability to perform hierarchical reasoning, effectively utilizing retrieved knowledge by pruning redundant or subordinate information.

\subsection{Instruction Prompts} \label{app:prompt}
This section presents the detailed prompt templates utilized throughout our experiments. 
Specifically, Figures~\ref{fig:judge_start}-\ref{fig:judge_end} illustrate the prompts used for the LLM judge evaluation. 
Figure~\ref{fig:inference} shows the prompt for answer generation ($\text{Instruct}_{\text{Gen}}$). 
Figures~\ref{fig:const_start}-\ref{fig:const_end} detail the prompts for meta-memory construction ($\text{Instruct}_{\text{Reflect}}$, $\text{Instruct}_{\text{Action}}$, and $\text{Instruct}_{\text{Filter}}$). 
Finally, Figure~\ref{fig:classification} presents the prompt used for the classification of meta-memory units.

\clearpage
\input{figure/prompt/prompt_judge_1}
\input{figure/prompt/prompt_judge_2}
\input{figure/prompt/prompt_judge_3}
\input{figure/prompt/prompt_judge_4}
\input{figure/prompt/prompt_judge_5}

\input{figure/prompt/prompt_infer}

\input{figure/prompt/prompt_summary}
\input{figure/prompt/prompt_operation}
\input{figure/prompt/prompt_update}

\input{figure/prompt/prompt_classify}

%% file: table/dataset.tex
\begin{table}[t]
\centering
\small
\begin{tabular}{lr}
\toprule
\textbf{Task} & \textbf{\# Examples} \\
\midrule
Single-Session User & 70 \\
Single-Session Assistant & 56 \\
Multi-Session & 133 \\
Temporal Reasoning & 133 \\
Knowledge Update & 78 \\
Single-Session Preference & 30 \\
\bottomrule
\end{tabular}
\caption{Statistics of the LongMemEval Benchmark.}
\label{tab:dataset}
\end{table}

%% file: table/infer_time.tex
\begin{table}[t]
\centering
\small
\begin{tabular}{lrr}
\toprule
\textbf{Method} & \textbf{Output Token} & \textbf{Inference Time (s)} \\
\midrule
\rowcolor{gray!8}\multicolumn{3}{l}{\textbf{\textit{Qwen3-30B-A3B-Instruct}}} \\
\midrule
Full Text & 252.72 & 10.95 \\
w/ MetaMem & 177.01 & 8.52 \\
\hdashline
RAG & 297.96 & 18.60 \\
w/ MetaMem & 339.53 & 21.05 \\
\hdashline
LightMem & 175.52 & 17.59 \\
w/ MetaMem & 184.72 & 19.87 \\
\midrule
\rowcolor{gray!8}\multicolumn{3}{l}{\textbf{\textit{Llama3.1-70B-Instruct}}} \\
\midrule
Full Text & 572.38 & 87.45 \\
w/ MetaMem & 497.31 & 74.83 \\
\hdashline
RAG & 80.89 & 10.77 \\
w/ MetaMem & 96.73 & 13.05 \\
\hdashline
LightMem & 84.80 & 11.23 \\
w/ MetaMem & 111.82 & 14.86 \\
\bottomrule
\end{tabular}
\caption{Inference Efficiency. We evaluate the average generated tokens and inference time of \method{} against several baselines. All experiments are conducted on the same device, with memory construction steps executed offline.}
\label{tab:infer_time}
\end{table}

%% file: table/k_fold.tex
\begin{table*}[t]
\centering
\small
\begin{tabular}{llccccccc}
\toprule
\multirow{2}{*}{\textbf{Fold Index}} & \multirow{2}{*}{\textbf{Method}} & \textbf{Single} & \textbf{Single} & \textbf{Multi} & \textbf{Temporal} & \textbf{Knowledge} & \textbf{Single} & \multirow{2}{*}{\textbf{Avg.}} \\
& & \textbf{User} & \textbf{Assistant} & \textbf{Session} & \textbf{Reasoning} & \textbf{Update} & \textbf{Preference} & \\
\midrule
\rowcolor{gray!9}\multicolumn{9}{l}{\textbf{\textit{Qwen3-30B-A3B-Instruct}}} \\
\midrule
\multirow{2}{*}{Fold 1} & LightMem & 100.0 & 37.5 & 61.1 & 51.5 & 82.6 & 100.0 & 67.7 \\
& \method{} & 91.7 & 37.5 & 55.6 & 60.6 & 91.3 & 100.0 & 70.7 \\ \hdashline
\multirow{2}{*}{Fold 2} & LightMem & 100.0 & 45.5 & 72.0 & 58.3 & 88.9 & 71.4 & 68.7 \\
& \method{} & 90.9 & 54.5 & 88.0 & 66.7 & 88.9 & 100.0 & 77.8 \\ \hdashline
\multirow{2}{*}{Fold 3} & LightMem & 100.0 & 33.3 & 65.6 & 75.0 & 72.2 & 83.3 & 69.7 \\
& \method{} & 90.9 & 33.3 & 78.1 & 80.0 & 83.3 & 83.3 & 75.8 \\ \hdashline
\multirow{2}{*}{Fold 4} & LightMem & 93.8 & 50.0 & 68.0 & 36.0 & 53.8 & 75.0 & 60.6 \\
& \method{} & 100.0 & 50.0 & 60.0 & 76.0 & 53.8 & 87.5 & 70.7 \\ \hdashline
\multirow{2}{*}{Fold 5} & LightMem & 75.0 & 7.7 & 64.5 & 70.6 & 71.4 & 100.0 & 62.6 \\
& \method{} & 80.0 & 15.4 & 64.5 & 64.7 & 78.6 & 100.0 & 64.6 \\
\midrule
\rowcolor{gray!9}\multicolumn{9}{l}{\textbf{\textit{Llama3.1-70B-Instruct}}} \\
\midrule
\multirow{2}{*}{Fold 1} & LightMem & 100.0 & 25.0 & 61.1 & 54.5 & 78.3 & 100.0 & 66.7 \\
& \method{} & 91.7 & 25.0 & 55.6 & 54.5 & 78.3 & 80.0 & 63.6 \\ \hdashline
\multirow{2}{*}{Fold 2} & LightMem & 90.9 & 27.3 & 84.0 & 58.3 & 88.9 & 42.9 & 66.7 \\
& \method{} & 100.0 & 45.5 & 80.0 & 69.4 & 77.8 & 71.4 & 73.7 \\ \hdashline
\multirow{2}{*}{Fold 3} & LightMem & 90.9 & 16.7 & 84.4 & 55.0 & 77.8 & 83.3 & 69.7 \\
& \method{} & 90.9 & 25.0 & 81.3 & 75.0 & 77.8 & 66.7 & 72.7 \\ \hdashline
\multirow{2}{*}{Fold 4} & LightMem & 100.0 & 33.3 & 76.0 & 36.0 & 84.6 & 75.0 & 65.7 \\
& \method{} & 93.8 & 25.0 & 80.0 & 56.0 & 69.2 & 62.5 & 66.7 \\ \hdashline
\multirow{2}{*}{Fold 5} & LightMem & 70.0 & 7.7 & 71.0 & 58.8 & 85.7 & 50.0 & 61.6 \\
& \method{} & 75.0 & 7.7 & 80.6 & 76.5 & 85.7 & 50.0 & 68.7 \\
\bottomrule
\end{tabular}
\caption{Performance of 5-Fold Cross-Validation between LightMem and \method{}.} \label{tab:k_fold}
\end{table*}

%% file: figure/unit_cnt.tex
\begin{figure}[t]
\centering
\subfigure[Qwen3-30B-A3B-Instruct.]{
\includegraphics[height=3cm]{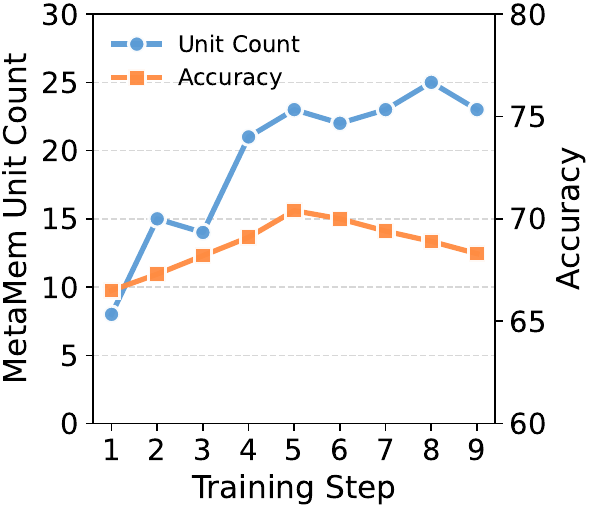}
\label{fig:cnt:a}
}
\subfigure[Llama3.1-70B-Instruct.]{
\includegraphics[height=3cm]{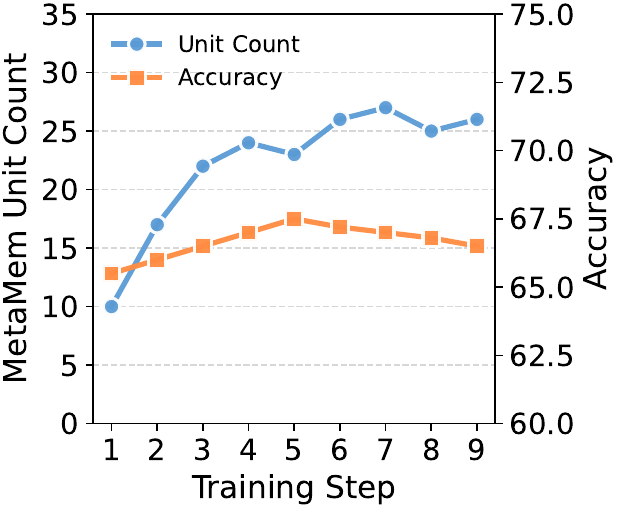}
\label{fig:cnt:b}
}
\caption{Evolution of Meta-Memory Unit Count and Accuracy across Training Steps.}
\label{fig:unit_count}
\end{figure}

%% file: table/judge_sensitivity.tex
\begin{table}[t]
\centering
\small
\resizebox{\linewidth}{!}{
\begin{tabular}{lr}
\toprule
\textbf{Method} & \textbf{Accuracy} \\
\midrule
LightMem & 67.5 \\
\hdashline
\method{} w/ Qwen3-235B-A22B Judge & 71.9 \\
\method{} w/ Qwen3-30B-A3B-Instruct Judge & 72.1 \\
\method{} w/ Llama3.1-70B-Instruct Judge & 71.6 \\
\bottomrule
\end{tabular}
}
\caption{Sensitivity Analysis of \method{} to Judge Model Choice during Meta-Memory Construction.}
\label{tab:judge_sensitivity}
\end{table}

%% file: table/case_study.tex
\begin{table*}[t]
\centering
\small
\renewcommand{\arraystretch}{1.5}
\resizebox{\linewidth}{!}{
\begin{tabular}{p{\linewidth}}
\toprule
\textit{\textbf{Case: Cumulative Numerical Reasoning: Road Trip Distance}} \\
\midrule
\textbf{Question:} What is the total distance I have driven across my four road trips?
\quad\quad \textbf{Ground Truth Answer:} 3000 miles \\
\hdashline
\textbf{Related Memories}: \\
\textit{2023/05/22 (Mon) 07:27} … charity run distances … \bad{5K} / \bad{10K} / marathon … \\
\textit{2023/05/23 (Tue) 16:14} … I've covered a total of \good{1,800} miles on my recent three road trips, including a solo trip to \good{Durango}, a weekend trip to \good{Breckenridge}, and a family trip to \good{Santa Fe} … \\
\textit{2023/05/23 (Tue) 16:14} … stopping at Maroon Lake … how long does it take … worth the detour … \\
\textit{2023/05/23 (Tue) 16:14} … planning a road trip from Denver to Aspen … scenic stops … \\
\textit{2023/05/26 (Fri) 14:27} … another road trip … just got back from an amazing 4-day trip to \good{Yellowstone National Park} … where we covered a total of \good{1,200 miles}. \\
\textit{2023/05/26 (Fri) 14:27} … thinking about our Yellowstone trip last month … we drove around \bad{300 miles} on the first day to reach Jackson, Wyoming … \\
\textit{2023/05/26 (Fri) 14:27} … planning road trip from Denver to Mount Rushmore … approximately \bad{570 miles} … \\
\hdashline
\textbf{Response of LightMem:} \\
Based on the conversation records, you have driven a total of \bad{2230} miles across four road trips. Below is a breakdown of the distances according to your own descriptions: \\
1. Yosemite National Park trip: \good{1,200} miles over four days. \\
2. Denver to Aspen: \bad{160 miles}. \\
3. Denver to Mount Rushmore: \bad{570 miles}. \\
4. California to Florida trip: only the first day (\bad{300 miles} to Jackson, Wyoming) was mentioned. \\
Summation: \good{1200} + \bad{160} + \bad{570} + \bad{300} = \bad{2230 miles}. \\
The initial statement that three trips totaled \good{1800 miles} introduces ambiguity, but since the California--Florida trip lacks a full distance, only explicitly stated values are used. Therefore, the computed total is \bad{2230 miles}.
\\
\hdashline
\textbf{Response of MetaMem:} \\
The total distance is obtained by summing explicitly stated trips: \\
(1) Three trips (\good{Durango, Breckenridge, Santa Fe}) explicitly stated to total \good{1800 miles}. \\
(2) A separate four-day \good{Yellowstone National Park} trip with \good{1200 miles}. \\
The mention of driving ``\bad{300 miles} on the first day'' is a partial segment already included in the 1200-mile Yellowstone total, so it is not added separately. Similarly, the Mount Rushmore trip is a future plan, not a completed trip. \\
Thus, the total distance is \good{1800 + 1200 = 3000 miles}. \\
\hdashline
\textbf{Used Meta-Memory Unit:} \\
(1) When performing numerical aggregation, prioritize explicit high-level summaries or total statistics provided by the user over fragmented partial details (e.g., daily breakdowns or single segments). \\
(2) Identify inclusion relationships to avoid double-counting subsets that are already encompassed within a reported total.
\\
\bottomrule
\end{tabular}
}
\caption{Case Study Comparing LightMem and MetaMem. Correctly utilized information is highlighted in \good{green}, while misleading or erroneous elements are highlighted in \bad{orange}.}
\label{tab:case_study}
\end{table*}

%% file: figure/prompt/prompt_judge_1.tex
\begin{figure*}[!h]
\centering
\includegraphics[width=\linewidth]{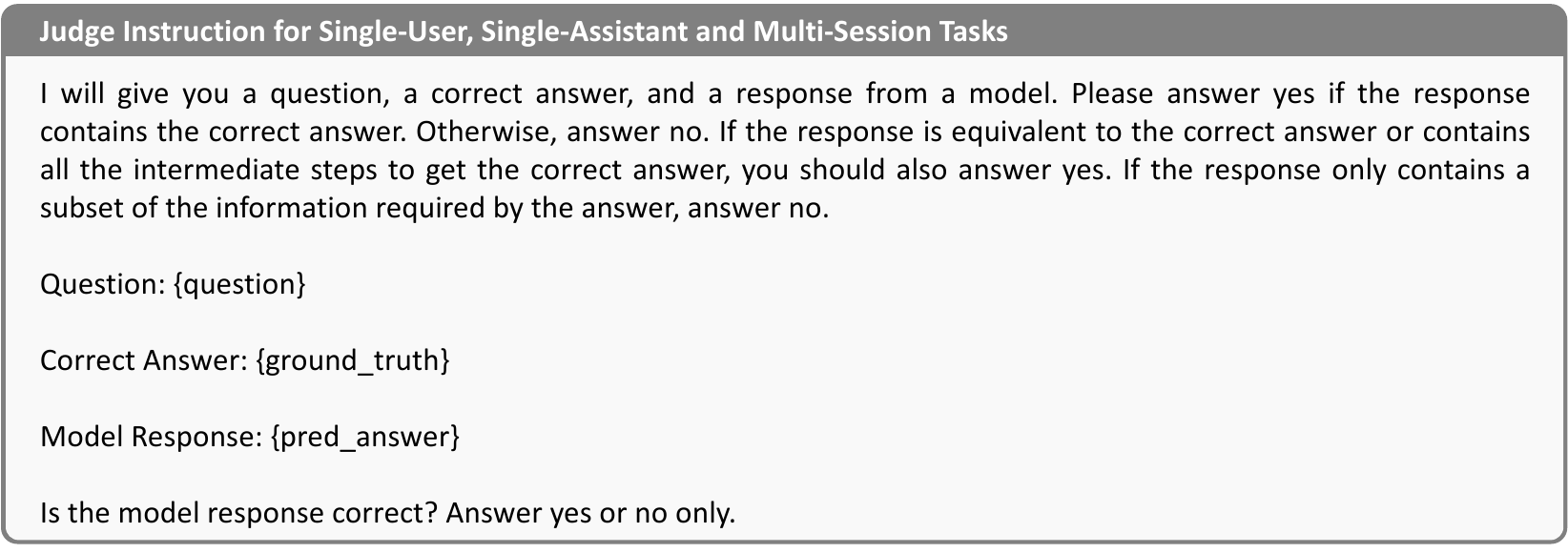}
\caption{Judge Instruction for Single-User, Single-Assistant and Multi-Session Tasks.}
\label{fig:judge_start}
\end{figure*}

%% file: figure/prompt/prompt_judge_2.tex
\begin{figure*}[h]
\centering
\includegraphics[width=\linewidth]{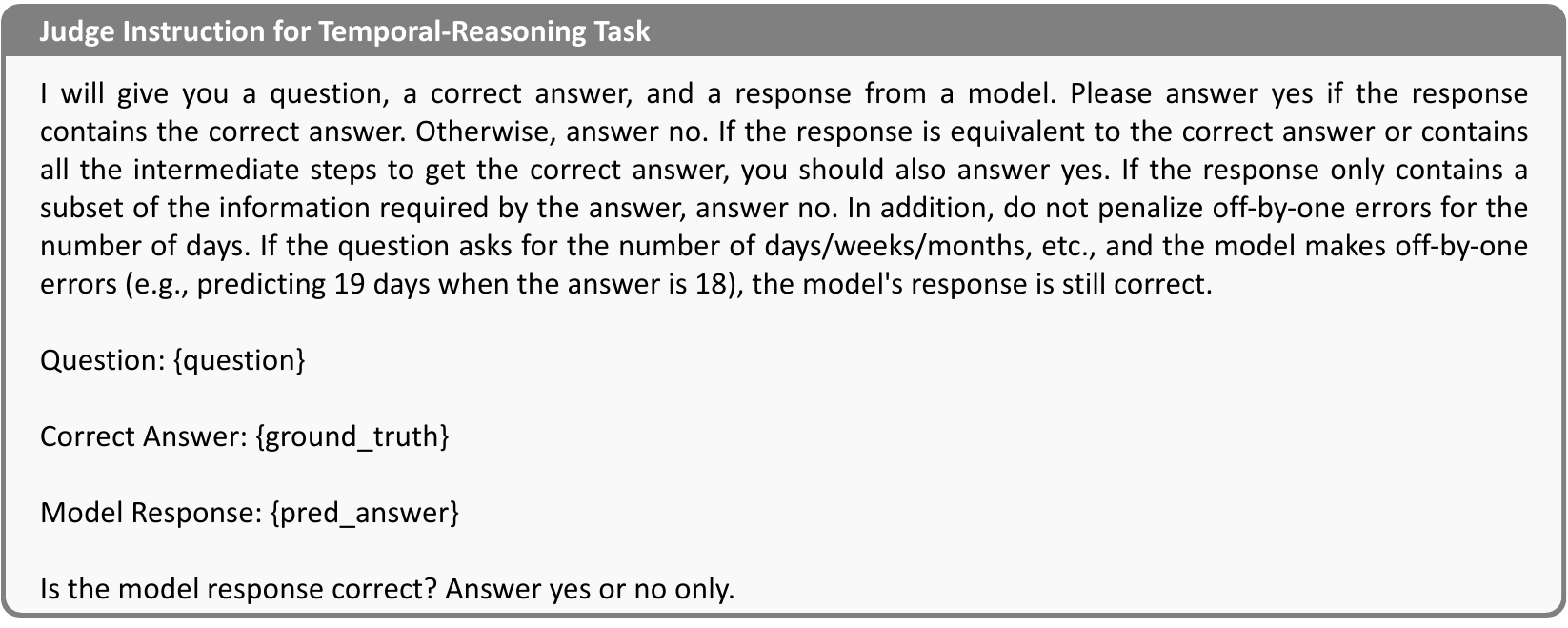}
\caption{Judge Instruction for Temporal-Reasoning Task.}
\end{figure*}

%% file: figure/prompt/prompt_judge_3.tex
\begin{figure*}[h]
\centering
\includegraphics[width=\linewidth]{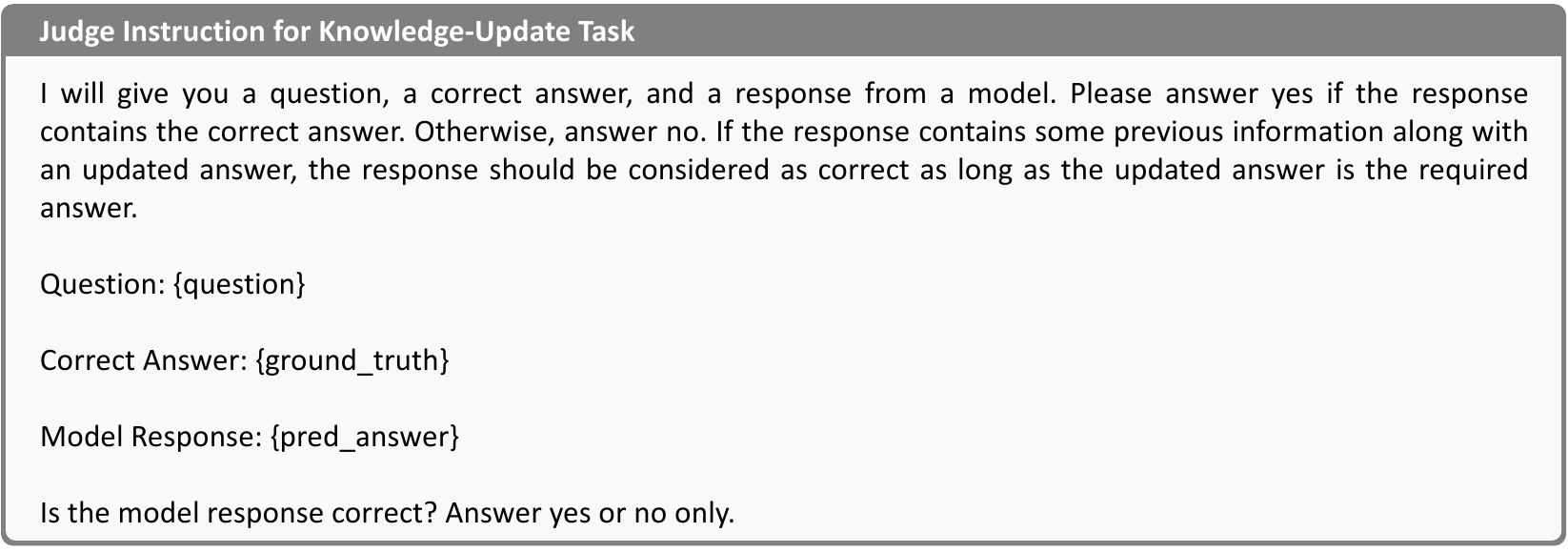}
\caption{Judge Instruction for Knowledge-Update Task.}
\end{figure*}

%% file: figure/prompt/prompt_judge_4.tex
\begin{figure*}[h]
\centering
\includegraphics[width=\linewidth]{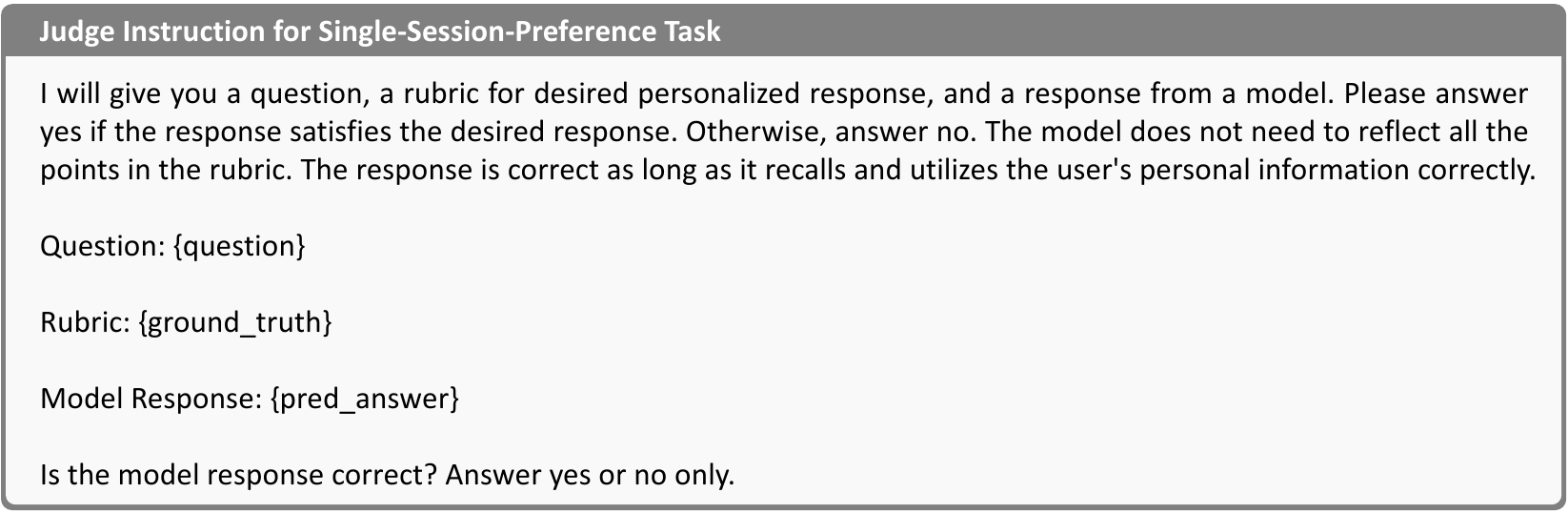}
\caption{Judge Instruction for Single-Session-Preference Task.}
\end{figure*}

%% file: figure/prompt/prompt_judge_5.tex
\begin{figure*}[h]
\centering
\includegraphics[width=\linewidth]{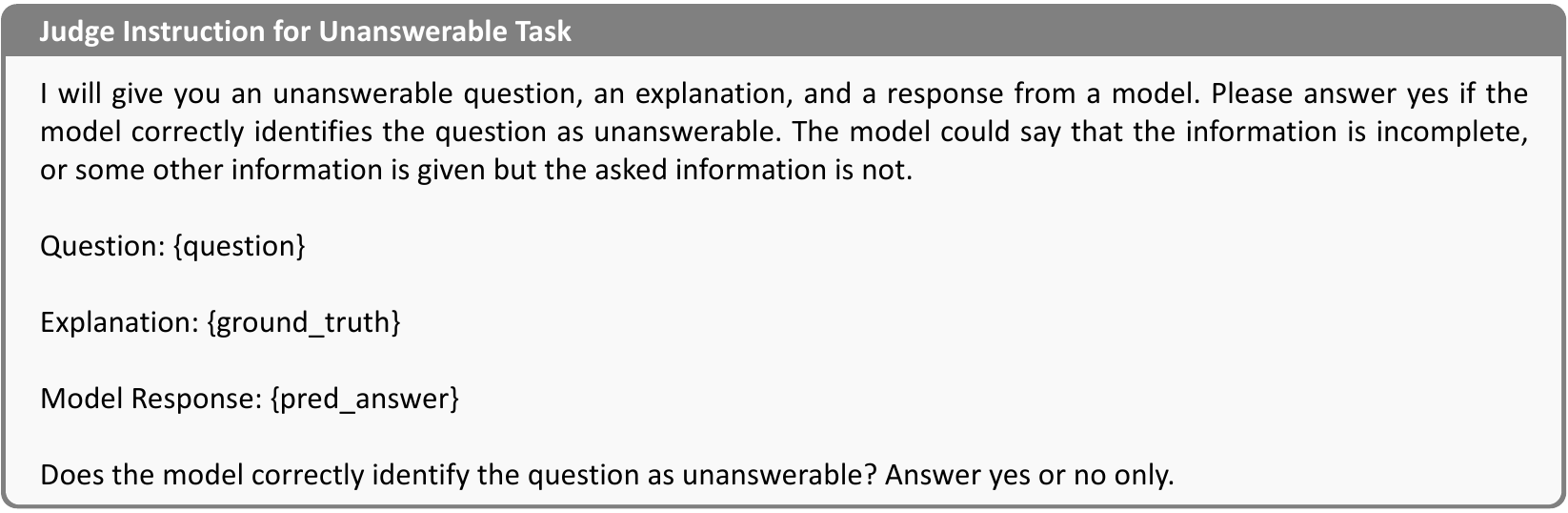}
\caption{Judge Instruction for Unanswerable Task.}
\label{fig:judge_end}
\end{figure*}

%% file: figure/prompt/prompt_infer.tex
\begin{figure*}[h]
\centering
\includegraphics[width=\linewidth]{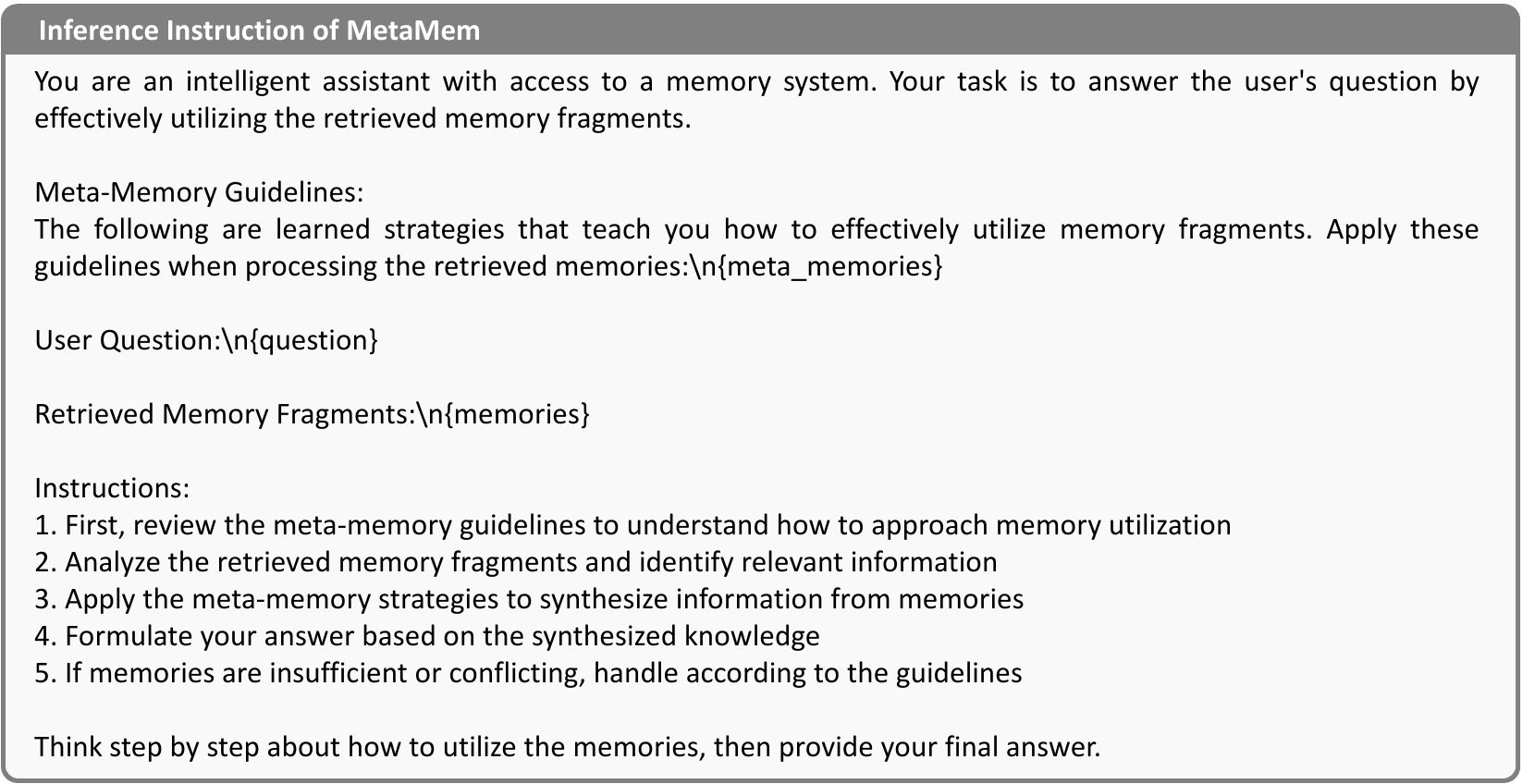}
\caption{Inference Instruction of MetaMem.}
\label{fig:inference}
\end{figure*}

%% file: figure/prompt/prompt_summary.tex
\begin{figure*}[h]
\centering
\includegraphics[width=\linewidth]{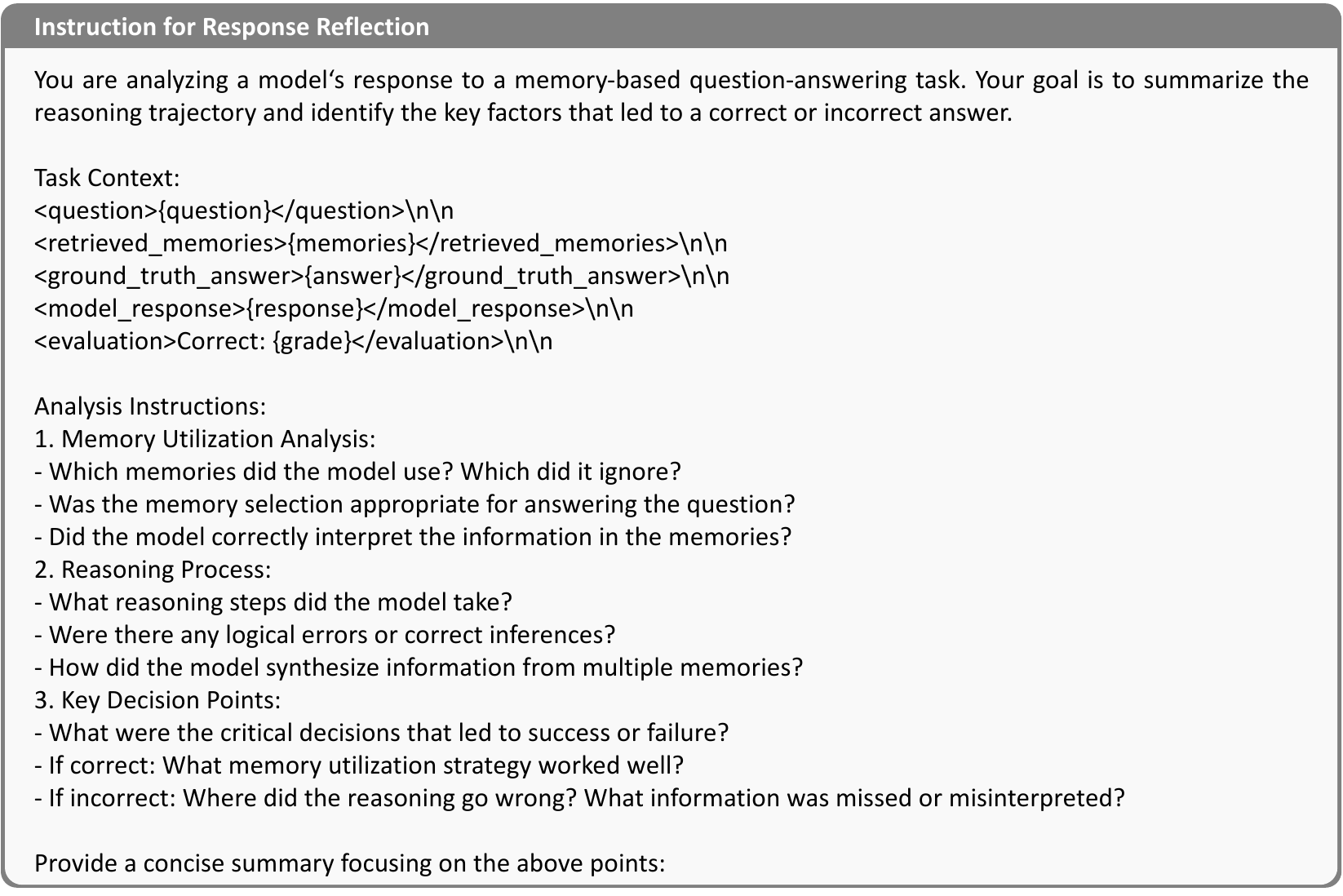}
\caption{Instruction for Response Reflection.}
\label{fig:const_start}
\end{figure*}

%% file: figure/prompt/prompt_operation.tex
\begin{figure*}[h]
\centering
\includegraphics[width=\linewidth]{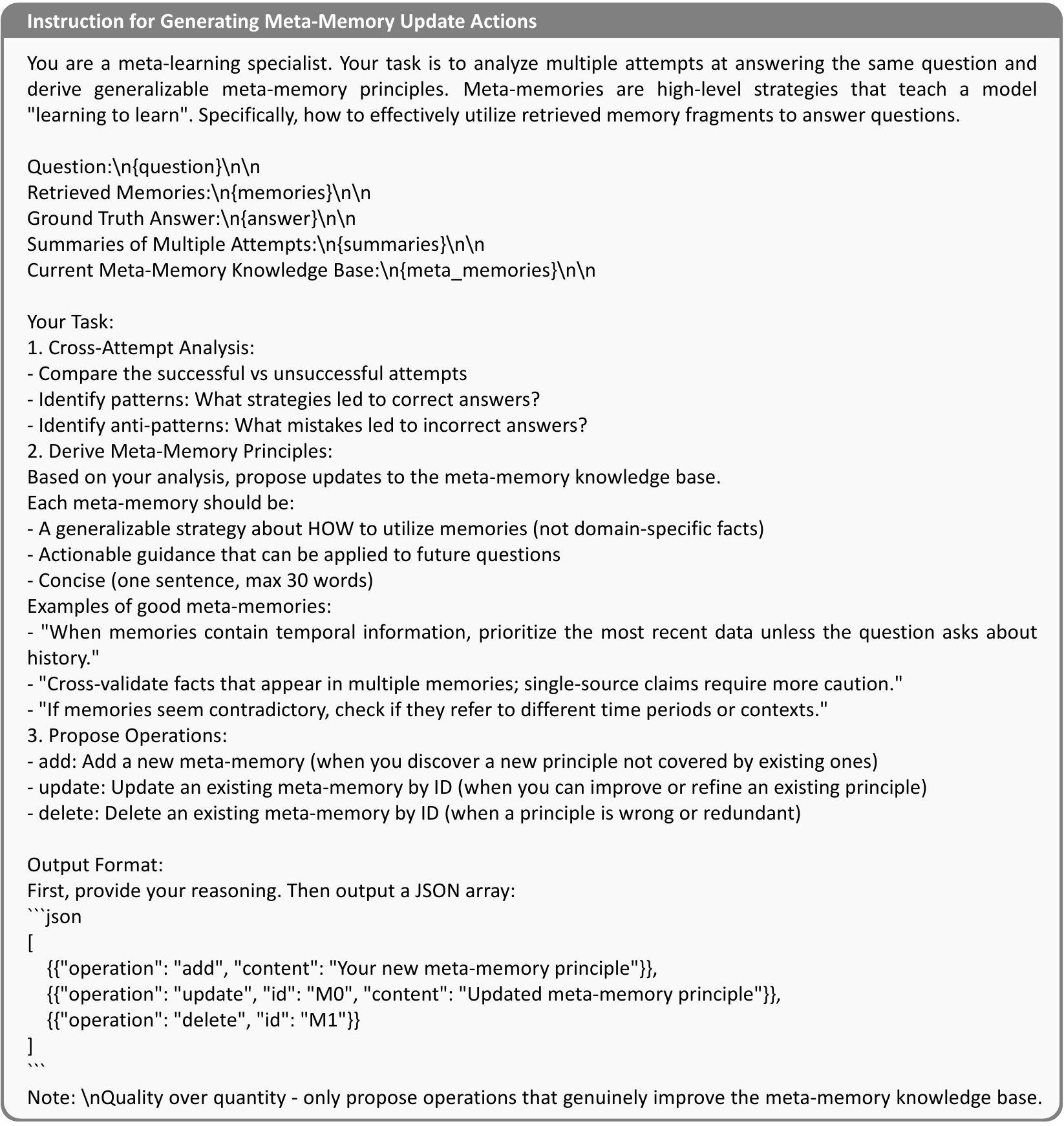}
\caption{Instruction for Generating Meta-Memory Update Actions.}
\end{figure*}

%% file: figure/prompt/prompt_update.tex
\begin{figure*}[h]
\centering
\includegraphics[width=\linewidth]{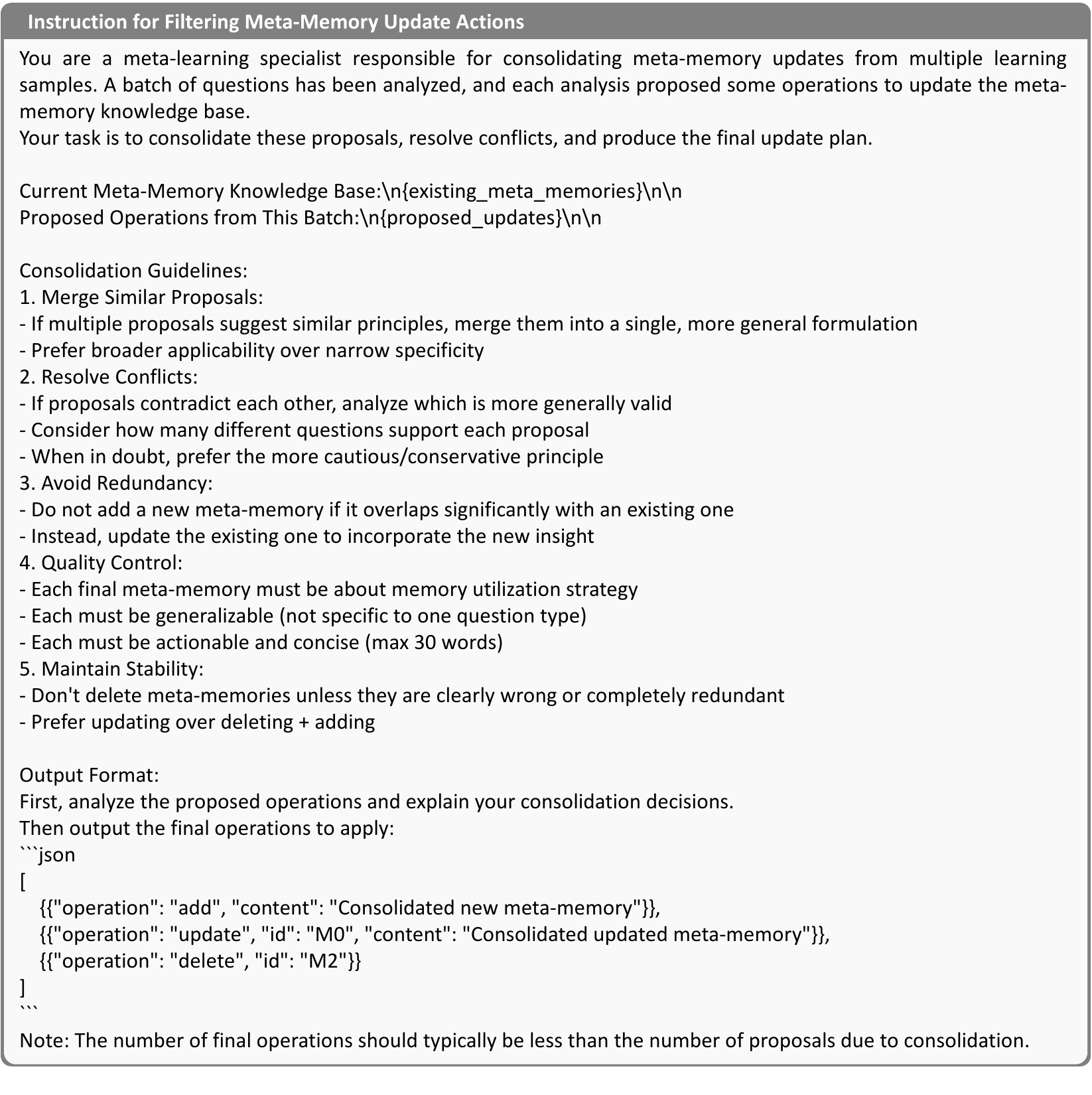}
\caption{Instruction for Filtering Meta-Memory Update Actions.}
\label{fig:const_end}
\end{figure*}

%% file: figure/prompt/prompt_classify.tex
\begin{figure*}[h]
\centering
\includegraphics[width=\linewidth]{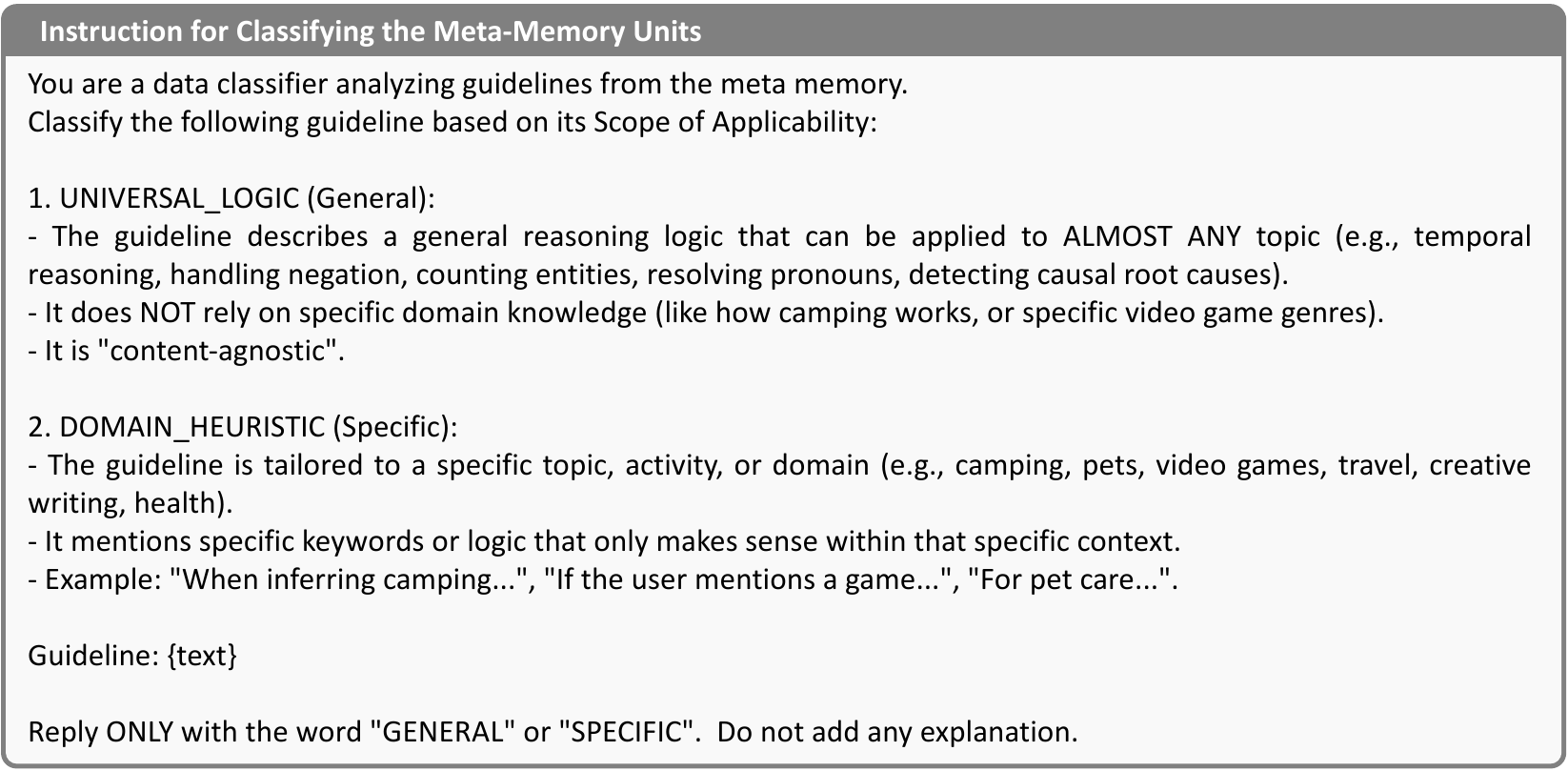}
\caption{Instruction for Classifying the Meta-Memory Units.}
\label{fig:classification}
\end{figure*}